\documentclass[11pt,a4paper]{article}
\usepackage[hyperref]{emnlp2020}
\usepackage{times}
\usepackage{latexsym}

\usepackage{microtype}

\aclfinalcopy 
\def\aclpaperid{1234} 

\usepackage{url}
\usepackage{graphicx}
\usepackage{subcaption}
\usepackage{caption}
\usepackage{comment}
\usepackage{textcomp}
\usepackage{amsthm, amsmath}
\usepackage{xcolor}
\usepackage{url}
\usepackage{times}
\usepackage{multirow}
\usepackage{adjustbox}
\usepackage{array}
\usepackage{amssymb}

\theoremstyle{definition}

\newcommand{\LAReQA}{LAReQA}

\newcommand{\EnEn}{\mbox{En-En}}
\newcommand{\XX}{\mbox{X-X}}
\newcommand{\XXmono}{\mbox{X-X-mono}}
\newcommand{\XY}{\mbox{X-Y}}
\newcommand{\TranslateEn}{\mbox{Translate-Test}}

\newcommand{\XLMR}{\mbox{XLM-R}}
\newcommand{\xquadr}{\mbox{XQuAD-R}}
\newcommand{\mlqar}{\mbox{MLQA-R}}

\newcolumntype{V}[2]{%
    >{\adjustbox{angle=#1,lap=\width-(#2)}\bgroup}%
    l%
    <{\egroup}%
}

\newcommand{\minus}{\scalebox{0.75}[1.0]{$-$}}

\title{LAReQA: Language-agnostic answer retrieval from a multilingual pool}

\author{
Uma Roy$^{\ddagger}$ \enskip Noah Constant \enskip Rami Al-Rfou \enskip Aditya Barua \enskip Aaron Phillips \enskip Yinfei Yang \\
Google Research
}
        
\date{}

\begin{document}
\maketitle
\begin{abstract}

We present LAReQA, a challenging new benchmark for language-agnostic answer retrieval from a multilingual candidate pool. Unlike previous cross-lingual tasks, LAReQA tests for ``strong'' cross-lingual alignment, requiring semantically related \emph{cross}-language pairs to be closer in representation space than unrelated \emph{same}-language pairs. Building on multilingual BERT (mBERT), we study different strategies for achieving strong alignment. We find that augmenting training data via machine translation is effective, and improves significantly over using mBERT out-of-the-box. Interestingly, the embedding baseline that performs the best on LAReQA falls short of competing baselines on zero-shot variants of our task that only target ``weak'' alignment. This finding underscores our claim that language-agnostic retrieval is a substantively new kind of cross-lingual evaluation.

\end{abstract}

\renewcommand{\thefootnote}{$^{*}$}
\footnotetext[1]{Corresponding authors: \\ \texttt{\{umaroy,nconstant\}@google.com}}
\renewcommand\thefootnote{\arabic{footnote}}

\renewcommand{\thefootnote}{$^{\ddagger}$}
\footnotetext[1]{Work done as a Google AI Resident.}
\renewcommand\thefootnote{\arabic{footnote}}

\section{Introduction}
\label{sec:intro}

Recent progress in self-supervised pretraining for language understanding has enabled training large multilingual models on 100+ languages at the same time, as in multilingual BERT (``mBERT'') and \XLMR{} \cite{BERT, XLM-R}. These models, despite being trained without any explicit objective of cross-lingual alignment, are surprisingly effective for cross-lingual transfer \cite{how-multi-is-mbert, surprising-xlingual-effectiveness, XLM-R}, suggesting that the models may have learned to ``factor out'' language and embed inputs into a language-agnostic space.

\begin{figure}[t]
\centering
\begin{subfigure}[b]{0.49\columnwidth}
    \centering
    \includegraphics[width=\columnwidth]{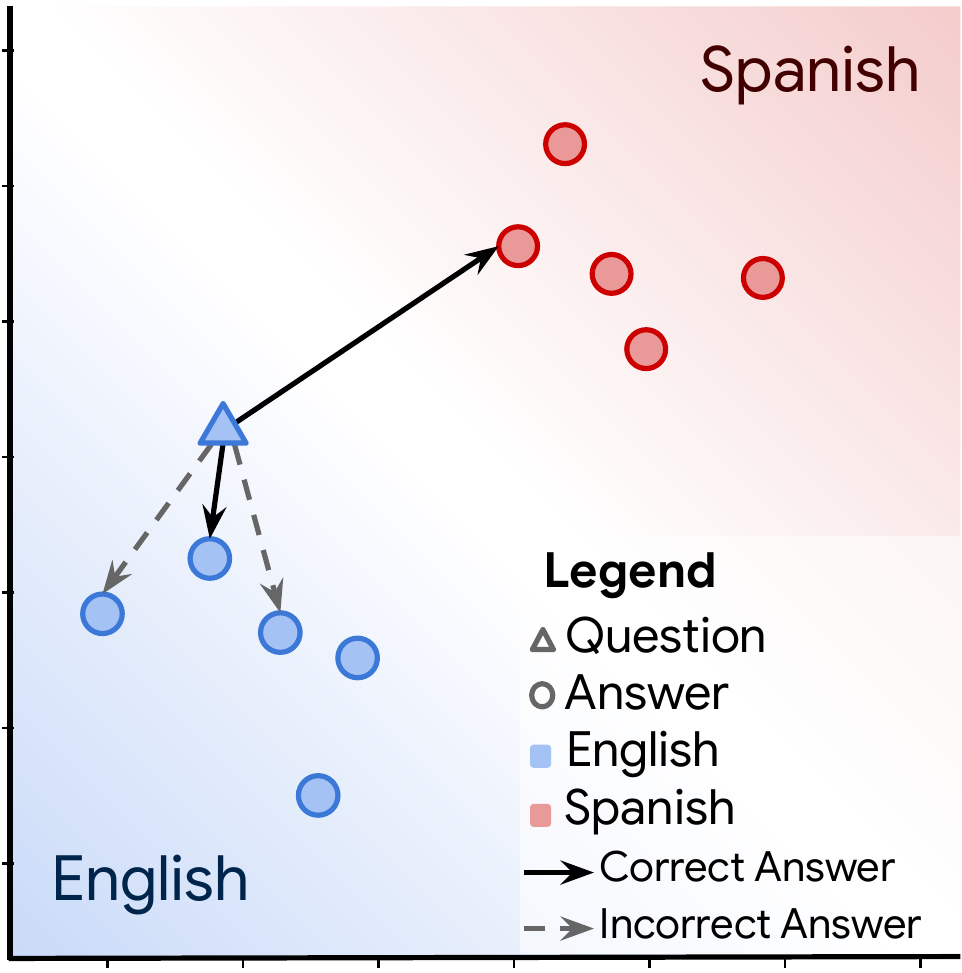}
    \caption{\textbf{Weak Alignment}}
    \label{fig:weak_alignment}
\end{subfigure}
\begin{subfigure}[b]{0.49\columnwidth}
    \centering
    \includegraphics[width=\columnwidth]{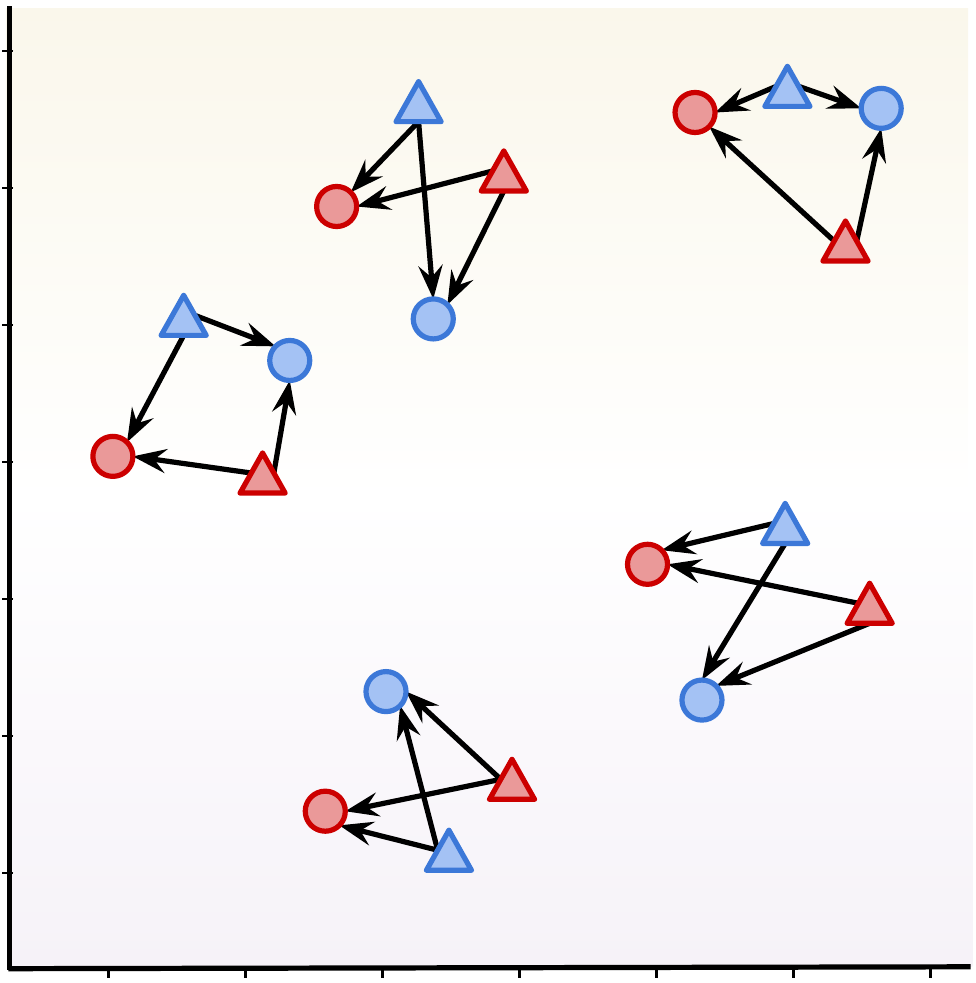}
    \caption{\textbf{Strong Alignment}}
    \label{fig:strong_alignment}
\end{subfigure}
\caption{A weakly aligned multilingual embedding space enables zero-shot transfer between languages, but incorrect answers in the same language are preferred over correct answers in a different language. A strongly aligned embedding space ``factors out'' language, so the most semantically relevant pairs are always the closest, regardless of language.}
\label{fig:crown}
\end{figure}

At the same time, \citet{surprising-xlingual-effectiveness} observe that mBERT representations at all layers are highly accurate ($>$96\%) at classifying language ID, which the authors find ``surprising given the model's zero-shot cross-lingual abilities''. This raises an interesting question. To what degree are models like mBERT and \XLMR{} language agnostic or easily adaptable to be so?
Have they effectively disentangled the language-specific signal from the underlying semantic content, with each occupying a separate subspace?  
Can the learned representations be adapted by a lightweight alignment procedure to be \emph{truly} language agnostic?

To address these questions, we introduce a challenging new task, LAReQA: Language Agnostic Retrieval Question Answering, which requires models to retrieve relevant cross-lingual answers from a multilingual candidate pool. Performing well on this task demands a stronger degree of cross-lingual alignment than previous cross-lingual evaluations like XNLI \cite{XNLI}. Concretely, we propose to distinguish ``weak'' vs.~``strong'' alignment, defined as follows and illustrated in Figures~\ref{fig:weak_alignment}--\ref{fig:strong_alignment}:

\paragraph{Weak Alignment} For any item in language $L_1$, the nearest neighbor in language $L_2$ is the most semantically relevant item. (The specific notion of ``relevance'' may vary across tasks.)

\paragraph{Strong Alignment} For any item, all semantically relevant items are closer than all irrelevant items, regardless of their language. Crucially, relevant items in different languages are closer than irrelevant items in the same language.\footnote{Stricter notions of cross-lingual alignment are possible, such as requiring that model representations remove any trace of the original text language, preventing language ID from being reconstructed. We treat these as sub-types of ``strong'' alignment, but leave their investigation for future work.}

\paragraph{} To our knowledge, LAReQA is the first cross-lingual benchmark to target strong alignment.\footnote{\citet{how-multi-is-mbert} develop a related heuristic by calculating the average vector delta between two languages, and testing how well translation targets can be retrieved by finding the closest neighbor to the source plus the delta.} Building on top of multilingual BERT, we develop and test several baseline models on LAReQA\@. We find that mBERT already exhibits strong alignment between \emph{some} language pairs, but that this can be improved significantly by leveraging machine translation to extend the set of training examples and encourage cross-lingual alignment. 

One observation that emerges from our experiments is that strong alignment comes at a cost. Specifically, our baseline that reaches the best LAReQA performance lags behind other baselines on the narrower task of retrieving relevant answers that match the question language.

Our main contributions are as follows: (1) We propose a new framework for classifying different degrees of cross-lingual alignment. (2) We propose a challenging new benchmark to evaluate language bias in representations, setting stricter notions of cross-lingual embedding space alignment. (3) We investigate the potential for multilingual BERT to achieve ``strong'' cross-lingual alignment, including various fine-tuning techniques to improve alignment. (4) We publish our trained models and LAReQA benchmarking code for others to reproduce.

\section{Looking for Answers across Languages}
\label{sec:task}

In this section, we present the task of \textbf{answer retrieval from a multilingual candidate pool}, and argue that this task goes beyond existing cross-lingual benchmarks in demanding models with strongly aligned multilingual representations. The task can be summarized as: given a question in one language and potential answers in many different languages, retrieve the best answer for the question, \emph{regardless} of language. We begin by describing why this task is both useful and challenging. Next, we compare this task with existing cross-lingual tasks, and show how they differ in their ability to measure ``language bias''.

\subsection{Retrieval from a Multilingual Pool}

Finding relevant answers to questions from a large multilingual candidate pool is not a contrived task. User-generated content on the web is increasingly multilingual\footnote{For example, see \url{https://en.wikipedia.org/wiki/Wikipedia:Size_of_Wikipedia}.}, and the best answer to a given question may be written in a different language than the question. If search engines were language agnostic, retrieved results would come from a wide range of languages. Of course, some results would have to be machine translated to be made interpretable to a given user. But in many cases, even a poorly translated relevant result is more helpful than a less relevant native result.

One domain where cross-lingual retrieval is particularly valuable is in searching over user-generated content such as reviews of products and businesses. For example, suppose a Thai speaker wants to know if a local library offers private meeting rooms, and this question is answered by an existing Arabic user review of the library. Being able to respond to the Thai question by surfacing (a translation of) the relevant Arabic review unlocks content that was previously inaccessible.\footnote{There are various options for how to implement such a cross-lingual retrieval system in practice, not all of which require a model to support cross-language matching. One solution would be to store English translations alongside results in the index, and translate all queries to English before performing search. Alternatively, one could pre-translate results into all languages ahead of time. However as these solutions require larger indices, we believe it is worth considering retrieving multilingual results directly with a cross-lingual model.}

\subsection{Language Bias}

From the modeling perspective, one of the main challenges in retrieving relevant answers from across languages is avoiding ``language bias'', where a model prefers one language over another. It's clear why this bias is harmful: if the model prefers answers in a given language, it is prone to retrieve \emph{irrelevant} results in that language over relevant results from another language.

The main type of language bias we observe in our experiments is \textbf{same-language bias}---the tendency for models to prefer answers that match the question language. This is illustrated in Figure~\ref{fig:weak_alignment}, where the embeddings cluster primarily by language, and incorrect same-language candidates are preferred over any cross-language candidate. For a multilingual model to avoid same-language bias, it must align text from different languages under a language-agnostic embedding space, as in \ref{fig:strong_alignment}.

\subsection{Taxonomy of Cross-lingual Tasks} 
\label{sec:taxonomy}

Existing cross-lingual tasks---including all tasks in the recent XTREME suite \cite{xtreme}---fall into two categories, as described below. Neither type allows us to diagnose language bias and test for language-agnostic embeddings. The key missing piece is that none of these tasks require the model to make a choice among candidates in different languages.

\paragraph{Monolingual Tasks in Many Languages} Most existing cross-lingual benchmarks are formed by collecting monolingual tasks across various languages. Often, these evaluations are framed in terms of zero-shot or few-shot transfer learning, with the assumption that a practitioner only has access to task-specific training data in a single language. For instance, XNLI \cite{XNLI} tests how well a model fine-tuned on an English classification task (natural language inference) can generalize to non-English versions of the same task. Similarly, \mbox{MLDoc} \cite{MLDoc} and \mbox{PAWS-X} \cite{PAWSX} test cross-language transfer on document classification and paraphrase identification respectively.

Several recent cross-lingual QA benchmarks can also be described as cross-lingual collections of monolingual tasks. For example, XQuAD \cite{XQuAD} extends the popular SQuAD \cite{SQuAD} benchmark to cover QA pairs in 11 languages, but models are only tested on finding answers in contexts that match the question language.\footnote{Due to its parallel construction, it is possible to construct ``mixed-language'' QA pairs from XQuAD\@. This is the approach we take in Section~\ref{sec:lareqa}.} TyDi~QA \cite{TyDiQA} and MLQA's \cite{MLQA} ``cross-lingual transfer'' task also fall under this category.

\paragraph{Cross-lingual Tasks with Monolingual Candidates} A second class of cross-lingual tasks tests whether a model can, given an input in language~X, identify a target in language~Y\@. However, crucially, the set of candidates is restricted to language~Y\@. Thus, while the task is inherently cross-lingual, it does not test for language bias.

BUCC \cite{BUCC} is a task of this type. Given an English sentence, the task is to retrieve the corresponding translation from a monolingual pool of candidates in another language. Similarly, Tatoeba \cite{LASER} tests retrieval of translation pairs between English and 112 languages, but is restricted to monolingual pools. Bilingual lexicon induction or BLI \cite{BLI} is a similar task of cross-lingual retrieval from a monolingual pool, but targeting words rather than sentences.\footnote{One could construct versions of BUCC, Tatoeba and BLI that test for strong alignment, by switching to multilingual candidate pools. It would be interesting to compare these benchmarks to LAReQA in future work. Note, however, that the resulting tasks are somewhat ``unnatural'', in that there is typically no need to consider same-language candidates when mining for translation pairs or building a bilingual lexicon.}

MLQA \cite{MLQA} is an extractive question answering task, and in one variant of the task, ``\emph{generalized} cross-lingual transfer'', the question and answer are drawn from different languages. However, even in this case, the candidate answers are restricted to spans within a specific (monolingual) paragraph of context, so there is no way to assess whether the model is biased in preferring answers in one language over another.

\section{LAReQA}
\label{sec:lareqa}

Having motivated the need for a cross-lingual benchmark that asks models to choose \emph{between} languages, we now present a concrete case of such a cross-lingual evaluation, \textbf{LAReQA: Language-Agnostic Retrieval Question Answering}.

\subsection{Constructing LAReQA}
\label{sec:construction}

Our goal is to construct a QA retrieval task over a large multilingual pool where many or most of the target answers are found in a different language than the question itself.  To achieve this, we take the existing cross-lingual \emph{extractive} QA tasks XQuAD and MLQA and convert them into \emph{retrieval} tasks: \textbf{\xquadr{}} and \textbf{\mlqar{}}. These sets are designed so as to include parallel QA pairs across languages, allowing us to match questions with answers from different languages.

XQuAD is constructed by taking 240 paragraphs from the SQuAD v1.1 dev set and professionally translating the questions and paragraphs into 10 languages. Thus each question appears in 11 different languages and has 11 parallel correct answers. MLQA is constructed by using LASER \cite{LASER} to mine parallel sentences from Wikipedia, which annotators then use to generate questions. Unlike XQuAD, the questions in MLQA have a variable number (2--4) of parallel correct answers across the corpus. Contexts surrounding the answer sentence are not guaranteed to be parallel. Additionally, MLQA only covers 7 of the 11 XQuAD languages. See \citet{XQuAD} and \citet{MLQA} for more details on these sets.

To convert these span-tagging tasks into retrieval tasks, we follow the procedure from ReQA \cite{ReQA}. Specifically, we break each contextual paragraph into sentences\footnote{We release \xquadr{} and \mlqar{}, annotated with sentence boundaries as generated by an internal sentence breaker. For Thai we use \url{https://pypi.org/project/thai-segmenter}.}, and include all sentences across the dataset as candidate answers. A sentence is considered a correct answer to a question if it contains the target answer span\footnote{For both XQuAD and MLQA, there were no cases where an answer span crossed a sentence boundary.} for either that question or an equivalent question in another language (as identified by \texttt{qas\_id}).\footnote{One sentence can be the correct answer for multiple questions (with different \texttt{qas\_id}), as long as it contains the relevant target answer spans. We include all sentences from contextual paragraphs as candidates, even those that do not answer any question.} Table~\ref{table:stats} shows the number of questions and candidates per language in \xquadr{} and \mlqar{}.\footnote{We use the MLQA dev set rather than the larger test set to keep the speed of evaluation reasonable.} While the contexts for XQuAD are parallel across languages, differences in sentence breaking lead to variations in the number of candidates per language.\footnote{Thai is an outlier, with around 70\% the sentences per paragraph as the other languages. This is likely due to erroneous or ambiguous sentence breaking. Note, Thai lacks explicit sentence boundary markers, and human agreement on sentence breaking is much lower than English \cite{thai-breaking}.}

\begin{table}
\setlength\tabcolsep{4pt} 
\begin{tabular}{l | r r || r r}
   & \multicolumn{2}{c||}{\xquadr{}} & \multicolumn{2}{c}{\mlqar{}}     \\
   & questions        & candidates        & questions        & candidates        \\ \hline
ar & 1190            & 1222             & 517             & 2545             \\
de & 1190            & 1276             & 512             & 2362             \\
el & 1190            & 1234             & -               & -                \\
en & 1190            & 1180             & 1148            & 6264             \\
es & 1190            & 1215             & 500             & 1787             \\
hi & 1190            & 1244             & 507             & 2426             \\
ru & 1190            & 1219             & -               & -                \\
th & 1190            & 852             & -               & -                \\
tr & 1190            & 1167             & -               & -                \\
vi & 1190            & 1209             & 511             & 2828             \\
zh & 1190            & 1196             & 504             & 2322             \\ \hline
\end{tabular}
\caption{Numbers of questions and candidates per language in \xquadr{} and \mlqar{}.}
\label{table:stats}
\end{table}

\subsection{Evaluation}
\label{sec:evaluation}

For our primary evaluation, we use the standard information retrieval metric ``mean average precision'' (mAP), which measures a model's ability to rank relevant results over irrelevant ones. This metric is suitable when there are multiple relevant results for a given query. In our case, an \xquadr{} question will have 11 relevant answers, while an \mlqar{} question will have 2--4 relevant answers.

Formally, given a set of questions $Q$ and a ranking function over all candidates, mean average precision is defined as in Equation~\ref{eq:map}, where $R_i$ is the number of correct answers for question $q_i$, $P@j(q_i)$ is the Precision@j for $q_i$ and $\textrm{rel}(i,\,j)$ is an ``indicator'' function with value 1 if the $j$-th ranked candidate for $q_i$ is correct, 0 otherwise.

\begingroup
\setlength{\abovedisplayskip}{0pt}
\setlength{\belowdisplayskip}{12pt}
\begin{equation}
\textrm{mAP} = \frac{1}{T} \sum_{q_i \in Q} \frac{1}{R_i} \sum_{j=1}^{K} P@j(q_i) \times \textrm{rel}(i,\,j)
\label{eq:map}
\end{equation}
\endgroup

The mAP metric falls between 0 and 1. Any model that ranks all $C$ correct answers in the top $C$ positions (regardless of order) will achieve a perfect 1.0. Note, such a model cannot have strong language bias, as it needs to rank correct answers in \emph{every} language highly. On the other hand, being free from language bias is not sufficient for high mAP\@. As a trivial example, a model that ranks candidates randomly will have a low mAP\@. In sum, performing well on LAReQA mAP requires both strong QA retrieval quality, as well as an absence of language bias.

\section{Baseline Models} \label{sec:baselines}

We consider several neural baseline models for evaluation on \LAReQA. All our baselines are ``dual encoder'' models \cite{gillick2018end}, encoding questions and candidate answers separately. Unlike full cross-attention models, this architecture enables retrieval through approximate nearest neighbor search, and thereby scales well to large-scale retrieval problems. For more discussion of dual encoders for deep retrieval, see \citet{gillick2018end} and \citet{ReQA}.


\begin{figure}
    \centering
    \includegraphics[width=\columnwidth]{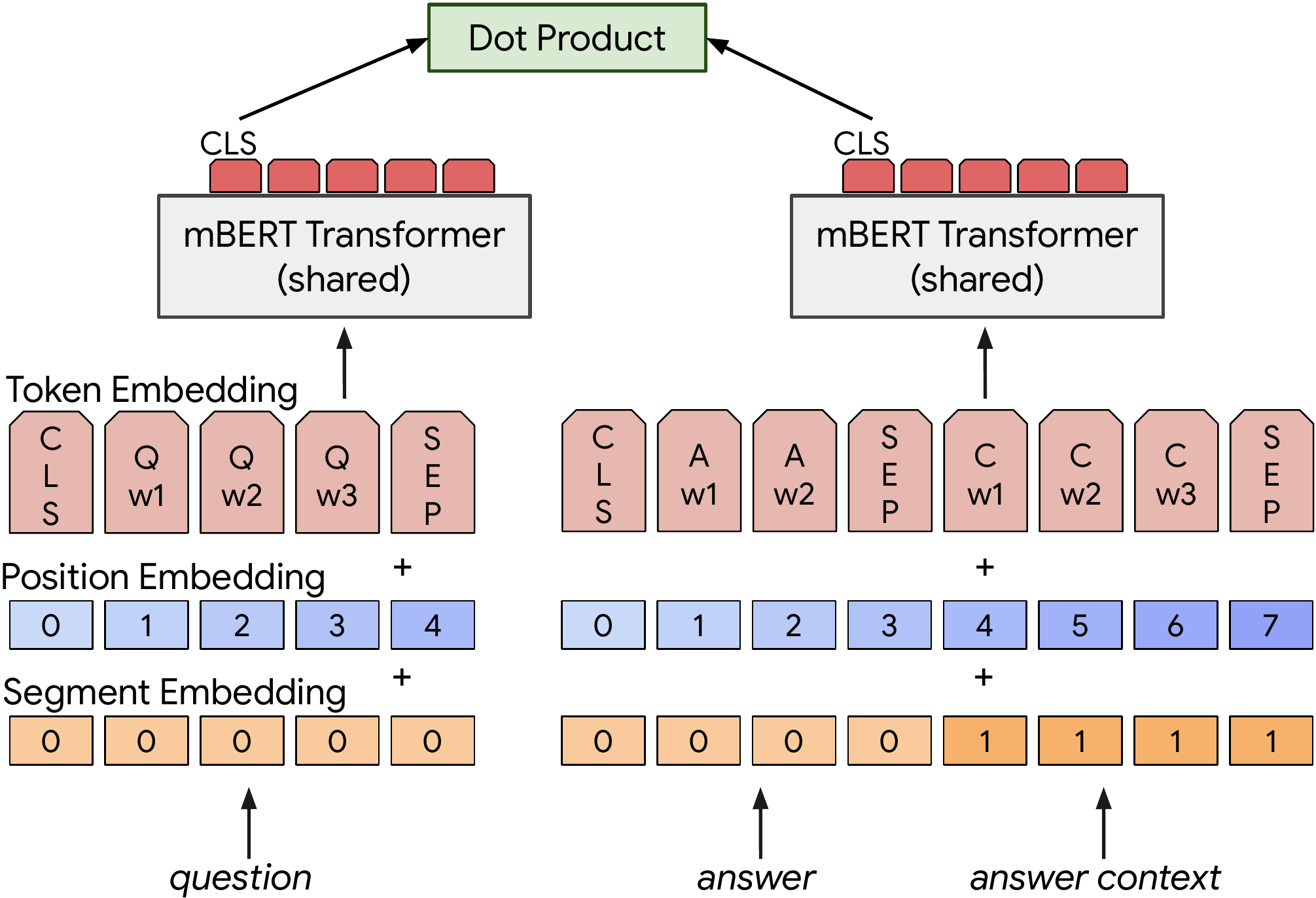}
    \caption{mBERT dual encoder architecture.}
    \label{fig:dualbert}
\end{figure}

Our baselines leverage multilingual BERT \cite{BERT}, or ``mBERT'', for cross-lingual pretraining. These baselines allow us to test (i)~how well mBERT already aligns languages in a language-agnostic space, and (ii)~the degree to which it can be adapted to do so.

We initialize each tower of a dual encoder model with pretrained mBERT, sharing weights between the question and answer encoding towers\footnote{When feeding inputs to the answer encoding tower, we concatenate the answer sentence and its containing context paragraph (``answer context"), using BERT's segment ids to distinguish between the two.}, as in Figure~\ref{fig:dualbert}. To obtain final question and answer encodings, we normalize the BERT CLS token to unit L2 norm. The model score for a QA pair $S(q,\,a)$ is the dot product of these encodings.

We fine-tune the mBERT towers for QA retrieval on SQuAD training data using an ``in-batch sampled softmax'' loss \cite{hendersonASSLGK17}, as this has been observed to converge quickly and perform well on retrieval tasks \cite{gillick2018end}.
The loss, given in Equation \ref{eq:rankloss}, encourages the score of a correct answer pair $S(q_i,\,a_i)$ to be higher than scores for incorrect answer pairings from the mini-batch $S(q_i,\,a_j)$:\footnote{In practice, we scale the similarity scores by a trainable constant factor before computing the softmax, as we observed this led to faster convergence.}

\begingroup
\setlength{\abovedisplayskip}{0pt}
\setlength{\belowdisplayskip}{12pt}
\begin{equation}
-\frac{1}{K} \sum_{i=1}^{K} \left[ S(q_i, a_i) - \mathrm{log} \!\!\! \sum_{j=1\,j\neq{}i}^{K} \!\!\! e^{S(q_i,\,a_j)} \right]
\label{eq:rankloss}
\end{equation}
\endgroup

\begin{figure}
\centering
\begin{subfigure}{.25\textwidth}
  \centering
  \includegraphics[width=.7\linewidth]{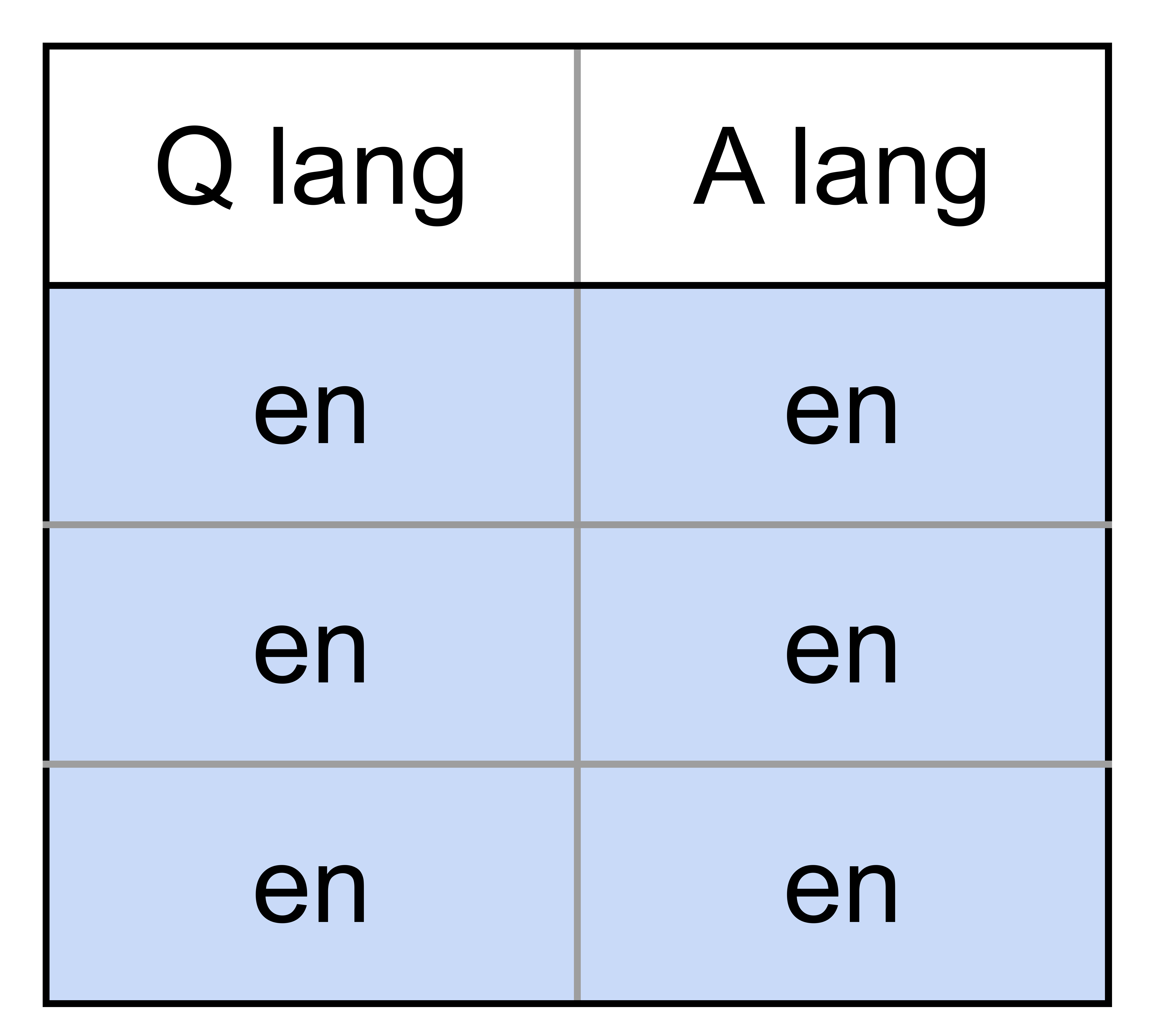}
  \caption{\EnEn{}}
  \label{fig:batch_en_en}
\end{subfigure}%
\begin{subfigure}{.25\textwidth}
  \centering
  \includegraphics[width=.7\linewidth]{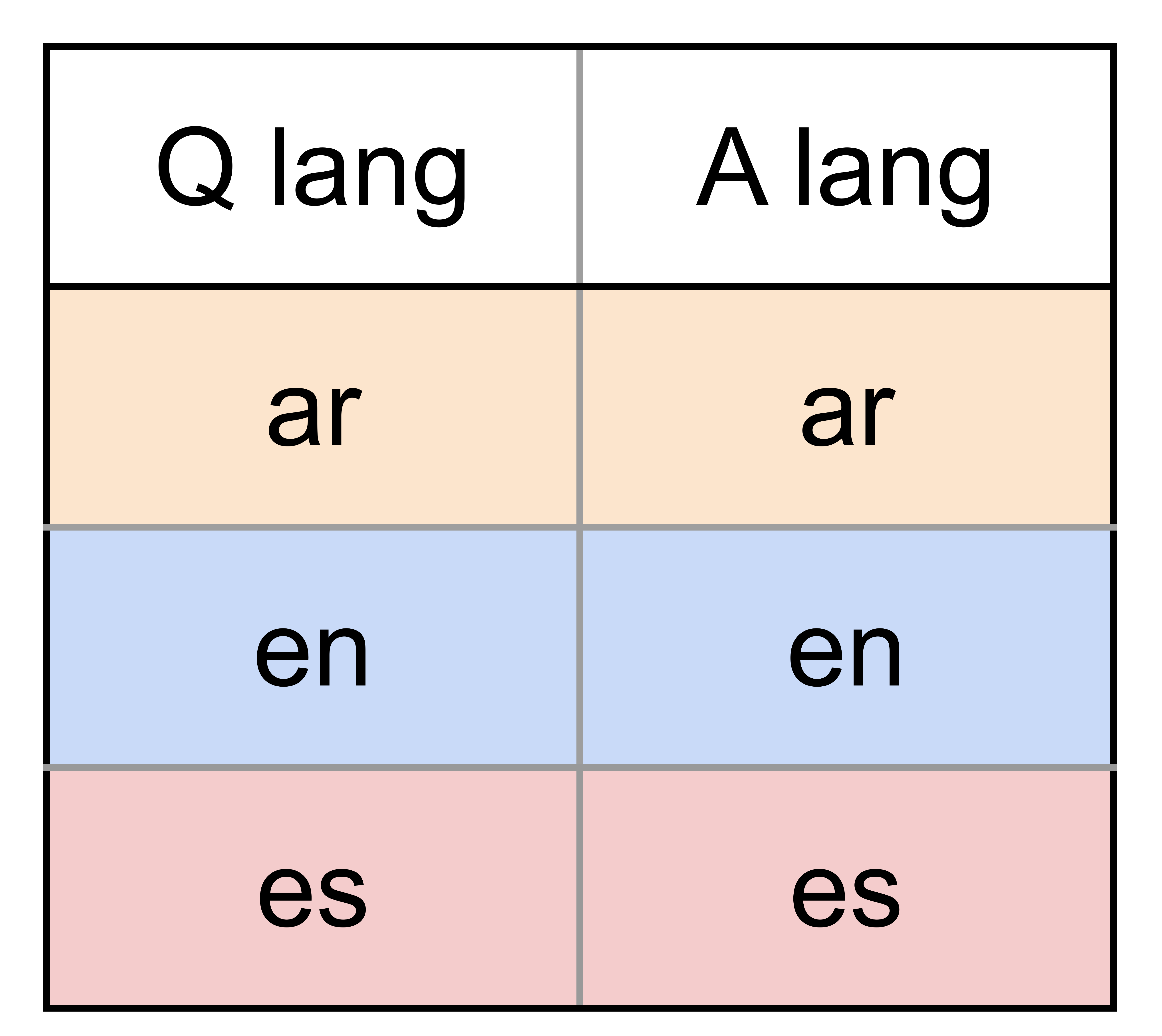}
  \caption{\XX{}}
  \label{fig:batch_X_X}
\end{subfigure}
\begin{subfigure}{.25\textwidth}
  \centering
  \includegraphics[width=.9\linewidth]{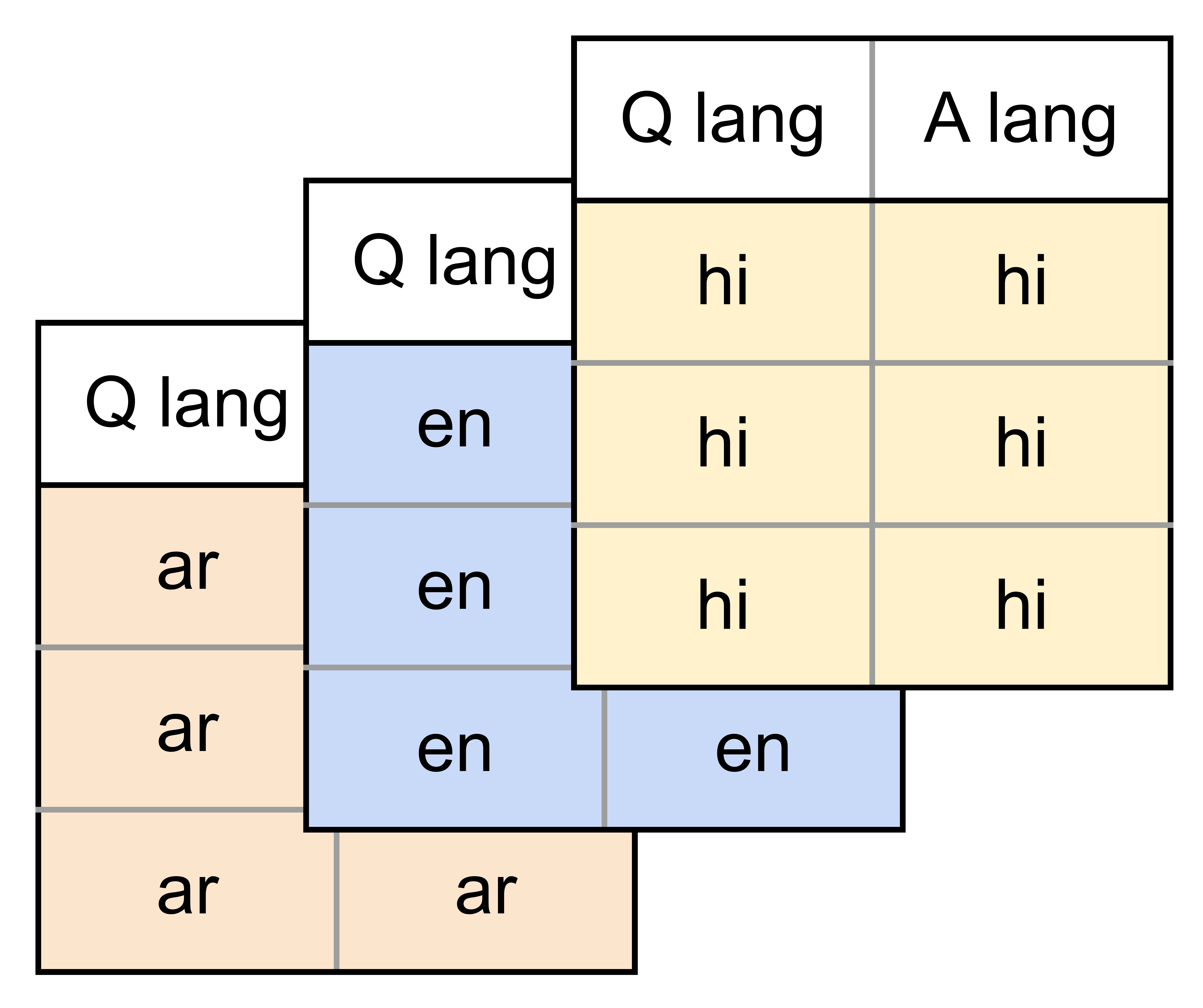}
  \caption{\XXmono{}}
  \label{fig:batch_X_X_mono}
\end{subfigure}%
\begin{subfigure}{.25\textwidth}
  \centering
  \vspace{1.1em}
  \includegraphics[width=.7\linewidth]{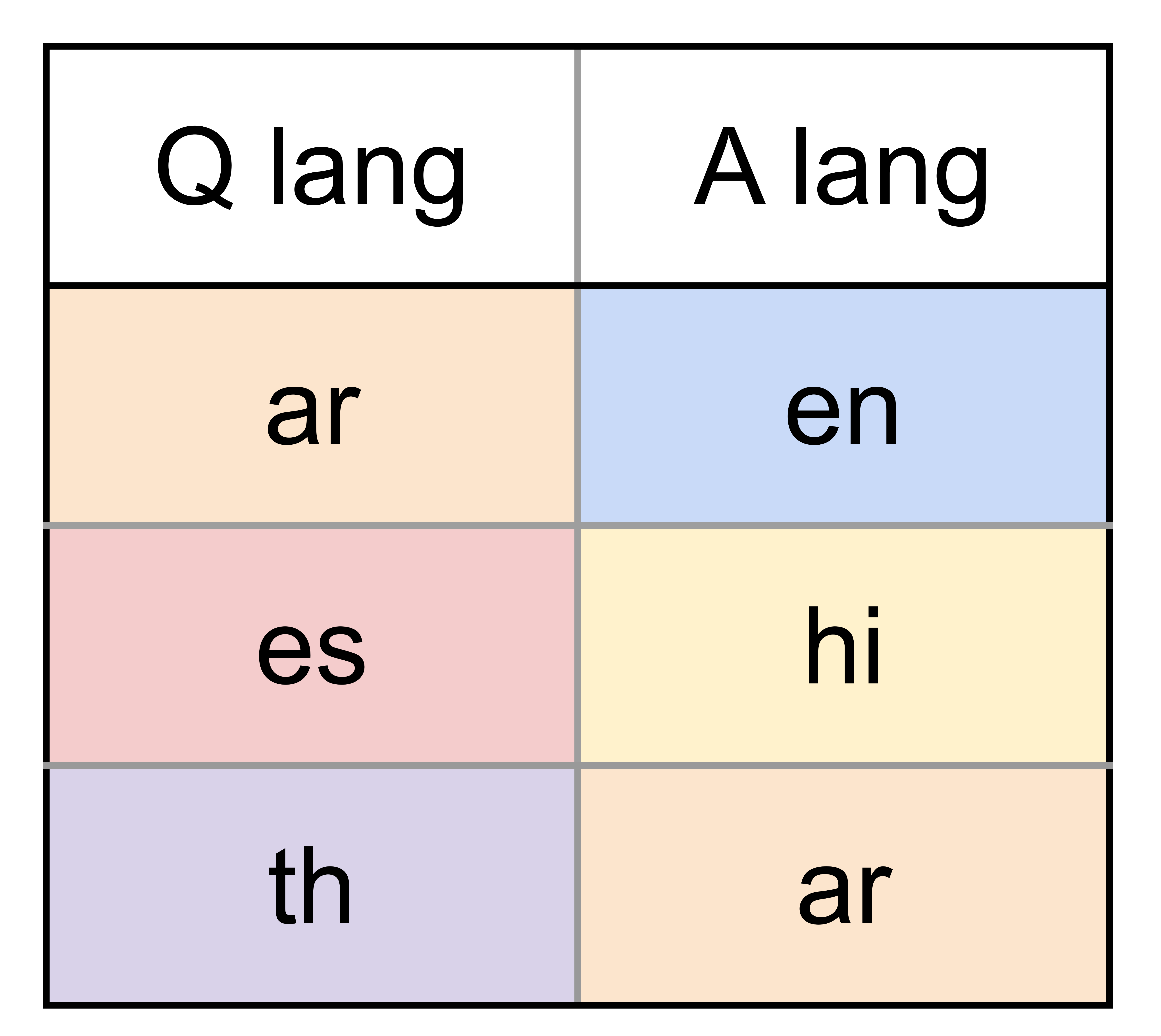}
  \caption{\XY{}}
  \label{fig:batch_X_Y}
\end{subfigure}
\caption{Sample batches for each baseline.}
\label{fig:batches}
\end{figure}

We train four variants of our mBERT model, using different fine-tuning datasets and batching procedures. Each model is fine-tuned on 32 TPU-v3 cores, with a batch size of 2,048. The in-batch sampled softmax is calculated separately per core, over sub-batches of 64 QA pairs. We use the standard BERT learning rate schedule, with an initial learning rate of 1e-5, which performed the best among \{1e-6, 3e-6, 1e-5, 3e-5\}. We train all baselines for 100,000 steps and observe no overfitting.

Our first baseline \textbf{``\EnEn{}''} adapts mBERT to QA retrieval by fine-tuning on the 80,000 English QA pairs from the SQuAD v1.1 train set, with the ranking loss from Equation \ref{eq:rankloss}. This baseline tests how well mBERT can perform language-agnostic retrieval when only fine-tuning on English data.

Our second baseline \textbf{``\XX{}''} extends the same SQuAD train set by translating each example into the 11 XQuAD languages using an in-house translation system. Within each example, the question and answer language are the same, giving 880,000 pairs total. If these pairs are shuffled and batched naively, as in Figure~\ref{fig:batch_X_X}, we expect the model to exhibit strong same-language bias, as all positive examples are within-language, while many in-batch negatives are cross-language. To avoid this bias, our third baseline \textbf{``\XXmono{}''} trains on the same examples, but ensures that each batch is monolingual, as shown in Figure~\ref{fig:batch_X_X_mono}.

Our fourth baseline \textbf{``\XY{}''} is similar to \XX{}, but allows a question and answer to be translated into different languages, giving 9,680,000 examples. This setup is the first to directly incentivize the model to treat answers from other languages as correct, which we expect to further reduce same-language bias.

Our final baseline \textbf{``\TranslateEn{}''} is not a proper text embedding model, as it relies on an external translation system at test time. Here, we simply translate any test data into English, and then score it with our \EnEn{} model.

Additionally, we compare the above baselines with the Universal Sentence Encoder Multilingual QA \cite{USE-QA}, which specifically targets cross-lingual QA retrieval. However, this model only supports 8 of the 11 XQuAD languages, and we found it was not competitive with our mBERT baselines, even when restricting the evaluations to the supported languages. See Appendix~\ref{sec:USE_QA} for details.

\section{Results and Analysis} 
\label{sec:results}

\subsection{LAReQA Performance}

We compare our five baseline models on the LAReQA task in Table~\ref{table:mAP}. On both \xquadr{} and \mlqar{}, the strongest model is the \TranslateEn{} baseline. This is unsurprising in that LAReQA demands language-agnostic retrieval, and \TranslateEn{} leverages an external machine translation system to actively ``remove'' the effects of language, by translating all test data to English.

\begin{table}
\setlength\tabcolsep{5pt} 
\begin{tabular}{l|rr}
                  & \xquadr{}  & \mlqar{} \\
\hline
En-En             & 0.29   & 0.36       \\
X-X               & 0.23   & 0.26   \\
X-X-mono          & 0.52   & 0.49    \\
X-Y               & 0.66  & 0.49    \\
Translate-Test    & 0.72  & 0.58  \\
\end{tabular}
\caption{Mean average precision (mAP) of baseline models on \xquadr{} and \mlqar{}\@.}
\label{table:mAP}
\end{table}%

Among the pure embedding approaches, \XY{} does the best on \xquadr{}, and is tied for best on \mlqar{}\@. The success of \XY{} shows that training directly on ``mixed-language'' QA pairs is helpful for the end task of language-agnostic retrieval from a multilingual pool.

As expected, \XXmono{} outperforms \XX{}, indicating that using a ranking loss with in-batch negatives is problematic when positives are within-language but negatives are mixed-language. Indeed, we will see shortly that \XX{} exhibits severe same-language bias.

For the remainder of the paper, we focus on \xquadr{}, as it is better balanced across languages than \mlqar{}, and the two sets showed similar results.

\subsection{Language Bias} \label{sec:lang-bias}

\begin{figure*}[htb!]
\centering
\begin{subfigure}{.35\textwidth}
  \centering
  \includegraphics[width=\linewidth]{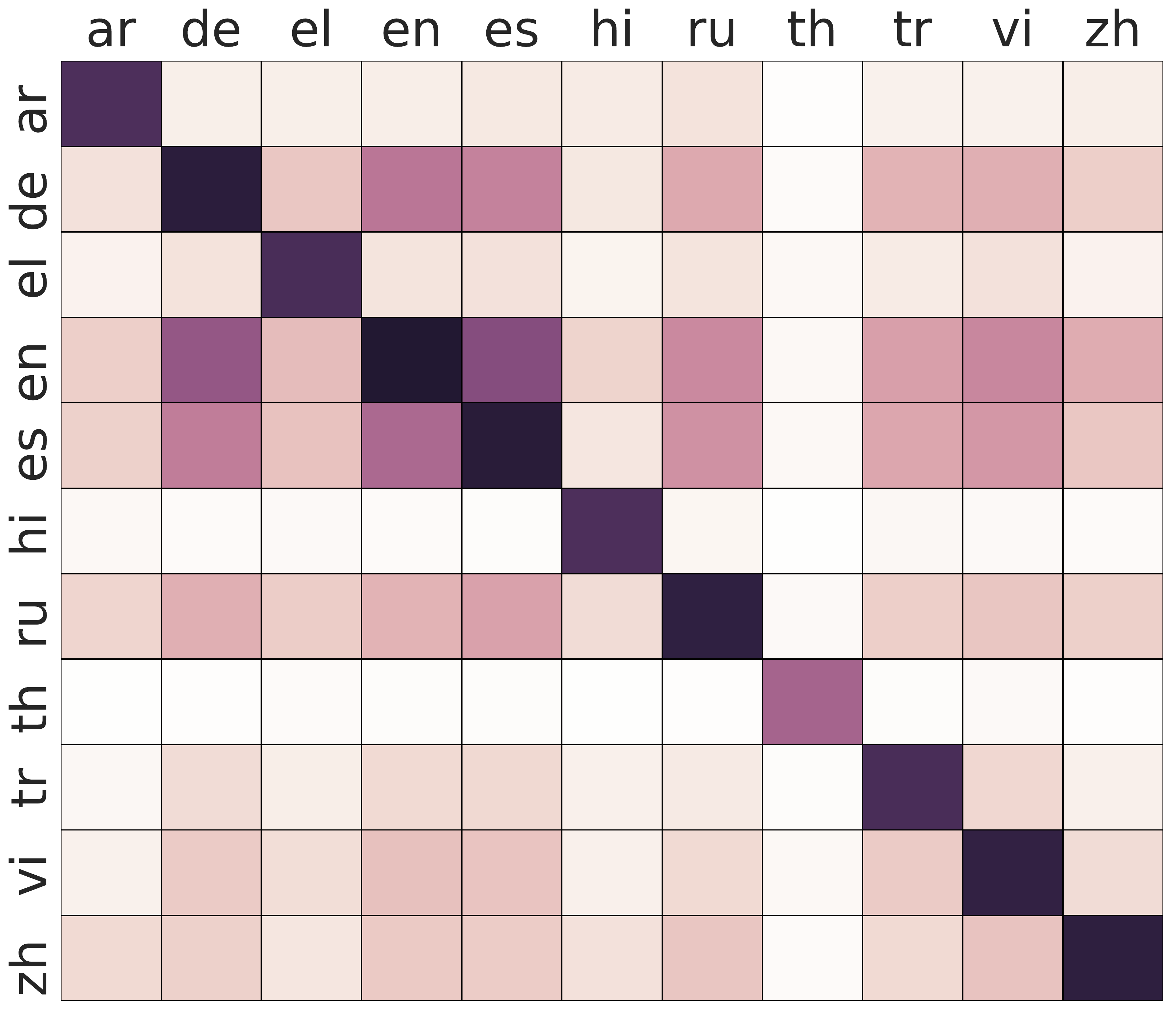}
  \caption{\EnEn{}}
  \label{fig:EnEn_single_ans}
\end{subfigure}%
\begin{subfigure}{.35\textwidth}
  \centering
  \includegraphics[width=\linewidth]{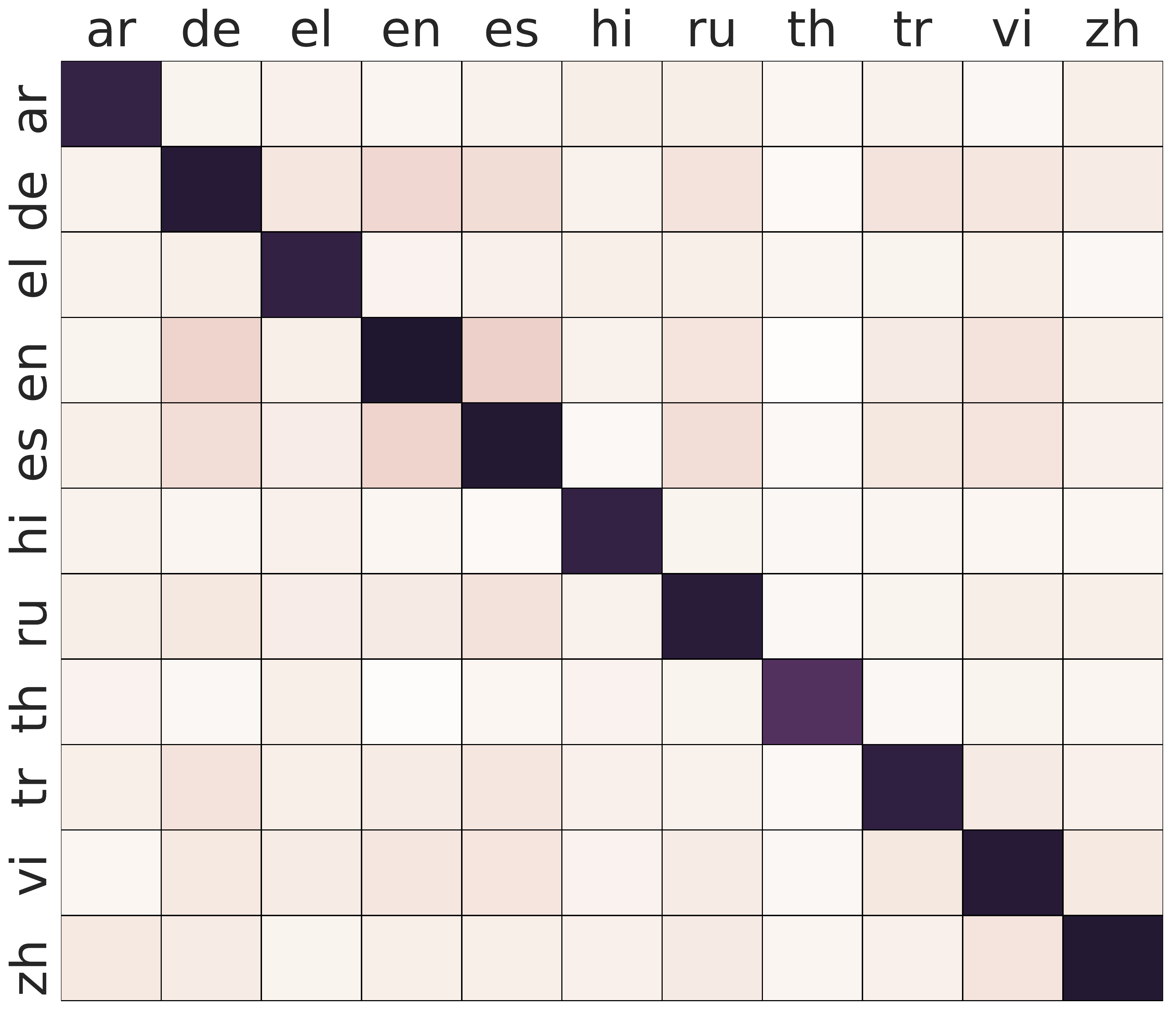}
  \caption{\XX{}}
  \label{fig:XX_single_ans}
\end{subfigure}
\begin{subfigure}{.048\textwidth}
  \centering
  \raisebox{12pt}{\includegraphics[width=\linewidth]{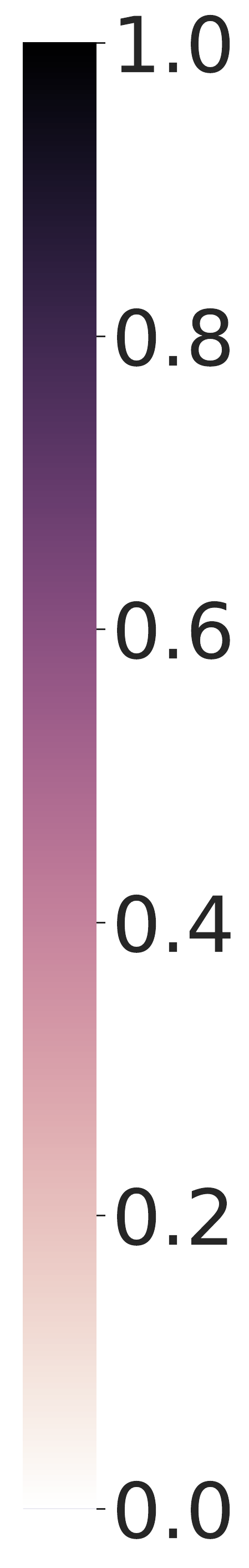}}
\end{subfigure}
\begin{subfigure}{.35\textwidth}
  \centering
  \includegraphics[width=\linewidth]{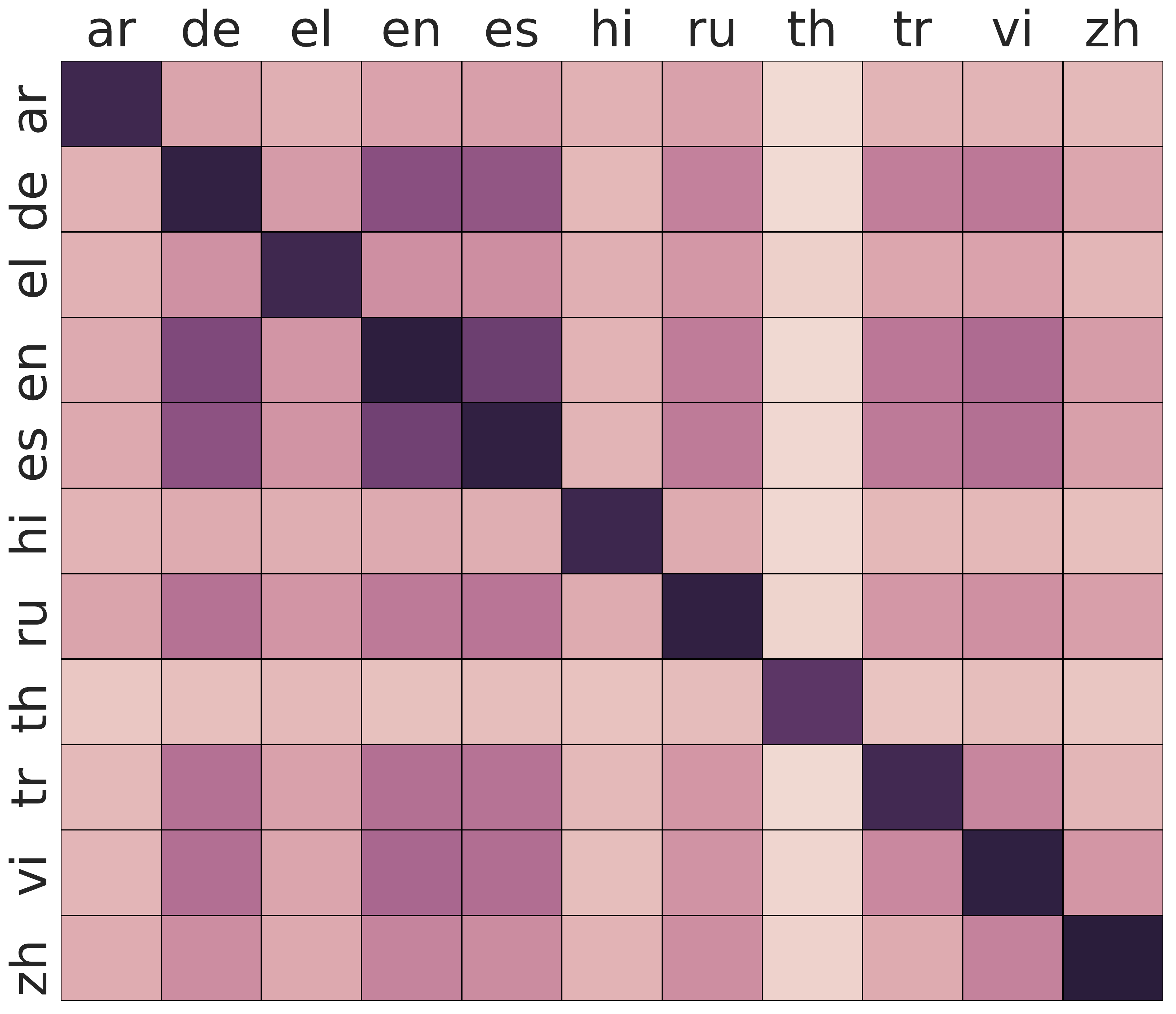}
  \caption{\XXmono{}}
  \label{fig:XX_mono_single_ans}
\end{subfigure}%
\begin{subfigure}{.35\textwidth}
  \centering
  \includegraphics[width=\linewidth]{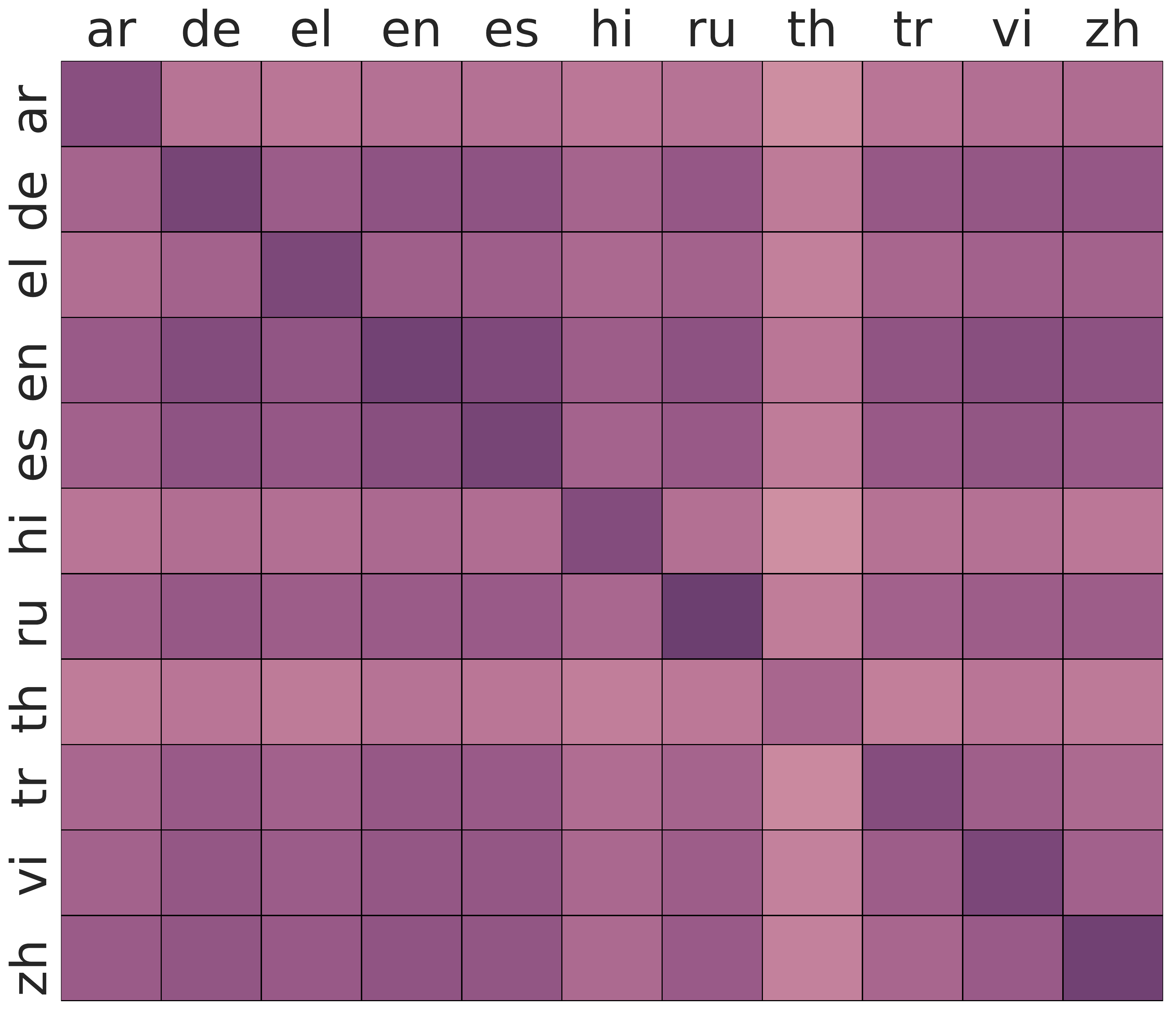}
  \caption{\XY{}}
  \label{fig:XY_single_ans}
\end{subfigure}
\begin{subfigure}{.048\textwidth}
  \centering
  \raisebox{12pt}{\includegraphics[width=\linewidth]{final_single_correct_ans/colorbar.pdf}}
\end{subfigure}
\caption{mAP on \xquadr{} broken down by question language (row) and answer language (column), when only one correct answer is included in the multilingual candidate pool.
}
\label{fig:single_ans}
\end{figure*}

We offer two additional analyses to more directly illustrate the language biases of our baselines. A third analysis, looking at the language distribution among top retrieved candidates, is given in Appendix~\ref{sec:top100_bias}, and is consistent with the results here.

\begin{table}

\setlength\tabcolsep{5pt} 
\begin{tabular}{l|rrrr}
                 & mAP  & mAP & & Rank \\
                &  \minus rand & \minus same & \% $\Delta$ & (\% $\Delta$) \\
\hline
En-En                     & 0.29                             & 0.22                     & 0.24          & 4        \\
X-X                       & 0.23                             & 0.15                     & 0.37          & 5   \\
X-X-mono                   & 0.52                             & 0.47                       & 0.10          & 2      \\
X-Y                       & 0.65                             & 0.64                      & 0.02         & 1        \\
Translate-Test            & 0.69                             & 0.60                       & 0.13          & 3            \\
\end{tabular}
\caption{Performance on modified versions of \xquadr{} where one target answer is removed, either from the same language as the question~($-$same) or a random other language~($-$rand).}
\label{tab:bias}
\end{table}%

\paragraph{Remove One Target} We rerun the \xquadr{} evaluation, but for each question, we remove one of its 11 target answers from the multilingual candidate pool. If a model is free of same-language bias, the effect of removing a single target should be constant, regardless of whether the removed target was in the same language as the question or not. Table~\ref{tab:bias} shows that in fact all our baselines perform better when a random \emph{cross-language} target is removed ($-$rand), as compared to the same-language target ($-$same). Looking at the delta between these conditions, the \XY{} baseline only displays a minimal bias, falling from 0.62 to 0.61 mAP\@. The most affected model is \XX{}, whose training procedure actively encouraged same-language bias. Interestingly even \EnEn{} shows a significant delta, indicating that simply fine-tuning mBERT on English QA is not sufficient to produce an embedding space that is strongly language agnostic.

\begin{table*}
\small
\setlength\tabcolsep{6pt} 
\begin{tabular}{l||l l l l l l l l l l l l || p{0.02\linewidth}p{0.02\linewidth}}

                  &     &     &     &     &     &     &     &     &     &     &     &   & \multicolumn{1}{V{90}{1em}}{Zero-shot} & \multicolumn{1}{V{90}{1em}}{LAReQA} \\
                  
                  & \textbf{ar}&
                    \textbf{de}&
                    \textbf{el}&
                    \textbf{en}&
                    \textbf{es}&
                    \textbf{hi}&
                    \textbf{ru}&
                    \textbf{th}&
                    \textbf{tr}&
                    \textbf{vi}&
                    \textbf{zh}&
                    \textbf{Avg}&
                    \multicolumn{2}{c}{\textbf{Rank}} \\
                    
                  \hline
En-En             & 0.76 & 0.87 & 0.78 & \textbf{0.90} & 0.87 & 0.76 & 0.85 & 0.51 & 0.77 & 0.84 & 0.85 & 0.80        & 4            & 4                   \\
X-X               & \textbf{0.83} & 0.87 & 0.84 & 0.89 & 0.88 & 0.83 & 0.86 & 0.75 & 0.84 & 0.87 & 0.88 & 0.85        & 2            & 5                   \\
X-X-mono          & \textbf{0.83} & \textbf{0.88} & 0.85 & \textbf{0.90} & \textbf{0.89} & \textbf{0.84}  & \textbf{0.87} & \textbf{0.76} & \textbf{0.85} & \textbf{0.88} & \textbf{0.89}       & \textbf{0.86}        & 1            & 3                   \\
X-Y               & 0.75 & 0.83 & 0.79 & 0.85 & 0.83 & 0.76 & 0.82 & 0.69 & 0.78 & 0.80 & 0.82 & 0.79        & 5            & 2                   \\
Translate-Test     & \textbf{0.83} & 0.87 & \textbf{0.86}  & \textbf{0.90} & 0.88 & 0.81 & 0.86 & 0.74 & 0.82 & 0.84 & 0.85 & 0.84        & 3            & 1                   \\
\end{tabular}
\caption{mAP on the zero-shot version of \xquadr{}, retrieving a single answer from a \emph{monolingual} pool that matches the question language. Note, only \EnEn{} is strictly ``zero-shot'', as the other models see machine translated fine-tuning data.} \label{table:zero_shot}
\end{table*}

\paragraph{Limit to One Target} As a more in-depth analysis of language bias, we evaluate on retrieval from a multilingual pool containing \emph{just one} correct answer. For \xquadr{}, since each question has 11 answers, this means evaluating on each target separately, with the other 10 targets removed from the pool. The heatmaps in Figure~\ref{fig:single_ans} show each baseline's mAP on single-answer retrieval, broken down by question language and answer language. Note, in this case mAP is equivalent to mean reciprocal rank (MRR)---the average inverse rank of the target answer. In line with our previous findings, all models display some degree of same-language bias, showing better performance on the diagonal, where Q and A languages match, than off-diagonal. The degree of bias matches the ranking from Table~\ref{tab:bias}. \XY{} displays the least bias, with most language pairs reaching over 0.4 mAP\@. As before, \XX{} is the most biased, but we also see significant bias in \EnEn{} and \XXmono{}\@.

These results also shed light on how well mBERT supports strong cross-lingual alignment ``out of the box''. Interestingly, even \EnEn{} shows fairly strong alignment among typologically related languages (e.g.~0.61 mAP on English-to-German and 0.57 on English-to-Spanish). These findings parallel those of \citet{how-multi-is-mbert} and \citet{surprising-xlingual-effectiveness}, who observe that mBERT zero-shot transfer is more effective among related languages. Our retrieval performance is lower between unrelated languages (e.g.~0.06 Arabic-to-Chinese), as well as on pairs where one of the languages is less well represented in mBERT's Wikipedia training data (Greek, Hindi and Thai).

While mBERT exhibits \emph{some} strong cross-lingual alignment out of the box, our results show that this can be improved greatly by using cross-lingual objectives, as in \XXmono{} and \XY{}\@. This finding echoes work by \citet{LASER}, \citet{XLM}, \citet{XLDA} and \citet{nmt_xlingual} showing that cross-lingual training can improve zero-shot transfer.

One point worth highlighting is the trade-off between on-diagonal and off-diagonal performance. If we limit attention to the diagonal, the models rank \XX{}\,$>$\,\XXmono{}\,$>$\,\EnEn{}\,$>$\,\XY{}. Thus, it appears there is a ``cost'' to strong cross-lingual alignment. For a given application, it may be worth sacrificing same-language quality to achieve better cross-language performance. However this begs the question: Is there any training technique that can achieve strong cross-lingual alignment without degrading within-language performance?

\subsection{Comparison to Standard Zero-Shot Cross-Lingual Transfer}

To highlight the difference between LAReQA and standard zero-shot cross-lingual evaluations like XNLI, we construct a zero-shot version of our QA retrieval task. We process the XQuAD data as before, but instead of a shared multilingual candidate pool, we restrict candidates to those matching the question language. Thus, like XNLI and the original XQuAD task, we're measuring a model's ability to generalize to monolingual tasks in new languages.

The performance of our baselines on this ``zero-shot'' retrieval from a monolingual pool is shown in Table~\ref{table:zero_shot}. Remarkably, the model ranking under this task diverges from that under our proposed LAReQA task of retrieval from a \emph{multilingual} pool. In particular, the \XX{}(-mono) baselines which were only trained on ``within-language'' examples now perform the best, beating the top LAReQA baselines \TranslateEn{} and \XY{}.

This result supports our claim that LAReQA tests for cross-lingual alignment in a way that existing zero-shot evaluations do not. Despite their strong language bias, visible in the dark diagonals in Figure~\ref{fig:single_ans}, the \XX{}(-mono) models give excellent performance in the typical zero-shot cross-lingual transfer scenario. Yet, as we saw in in Table~\ref{table:mAP}, these baselines are fundamentally ill-suited for retrieval from a multilingual pool, which demands strongly aligned multilingual embeddings. As an extreme case, \XX{} scored a mere 0.23 on LAReQA mAP, compared to the best embedding model \XY{} and the best overall baseline \TranslateEn{} with 0.63 and 0.70 respectively.

\subsection{Embedding Spaces}

\begin{figure}
    \centering
    \includegraphics[width=0.5\textwidth,trim={1em 0 0 0},clip]{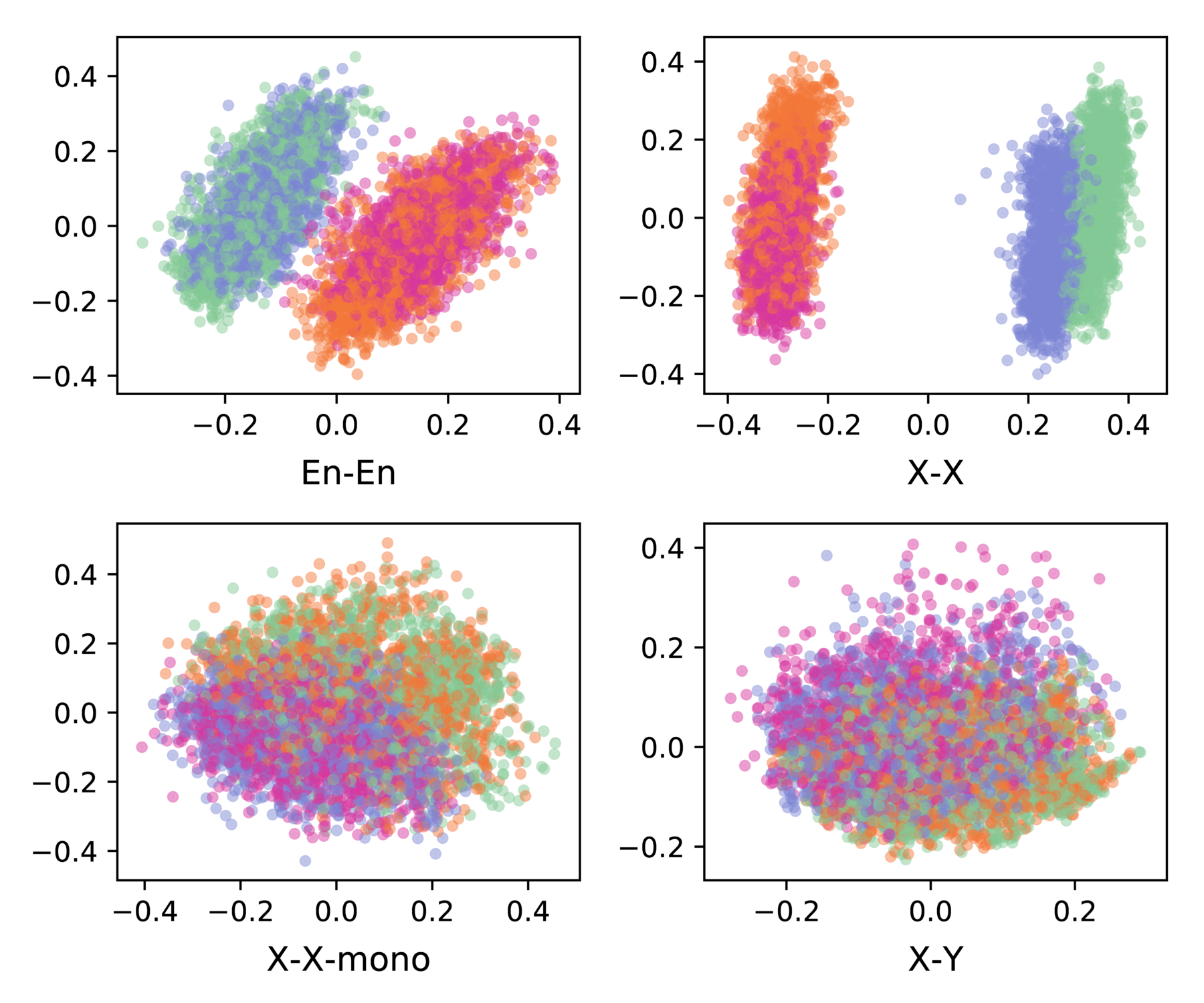}
    \includegraphics[width=0.49\textwidth,trim={0.27em 0 0 0},clip]{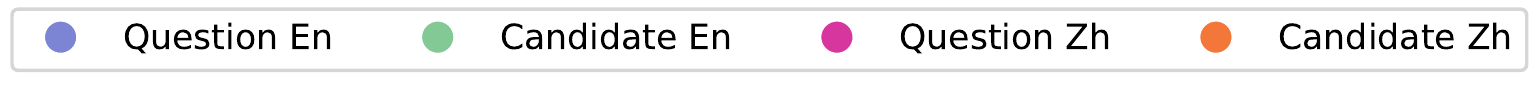}
    \caption{Model embeddings of all English and Chinese questions and candidates from \xquadr{}, visualized under 2D PCA.}
    \label{fig:embeddings}
\end{figure}

Figure~\ref{fig:embeddings} plots the first two principal components of each baseline's embeddings of all \xquadr{} questions and candidates in English and Chinese (chosen as they are distant languages). The \XX{} embeddings show a dramatic separation between Chinese and English. This is a clear case of weak alignment, where a model achieves good zero-shot performance (cf.~Table~\ref{table:zero_shot}) despite its embeddings being principally determined by language.

More generally, we observe that the more a model separates languages in embedding space, the worse it performs on LAReQA (cf.~Table~\ref{table:mAP}). This ordering is also reflected in the degree to which language ID is predictable from the embeddings. When we use a linear regression to predict language (English vs.~Chinese) from the question and candidate embeddings, the accuracies on a one-third holdout are \XX{}:~99.2\%, \EnEn{}:~97.7\%, \XXmono{}:~87.5\% and \XY{}: 54.0\%. This supports the claim that LAReQA is a better test of strong alignment than current zero-shot tasks.

\section{Conclusion}

LAReQA is a challenging new benchmark testing answer retrieval from a multilingual candidate pool. It goes further than previous cross-lingual benchmarks in requiring ``strong'' cross-lingual alignment, which is a step closer to truly language-agnostic representations.

We believe there is significant headroom for models to improve on LAReQA\@. Our best initial baseline sidesteps the alignment problem by simply translating all test data to English. Among embedding-based models, our strongest baseline (``\XY{}'') actively removes language bias by augmenting training data to include machine-translated cross-lingual examples. However, to achieve strong cross-lingual alignment, this model sacrifices performance on both retrieval from a monolingual pool (Table~\ref{table:zero_shot}), as well as retrieval of same-language candidates (Figure~\ref{fig:XY_single_ans} diagonal). It is an interesting question for future work whether strong alignment always comes at a cost, or if better training techniques will lead to models that can improve on all these measures simultaneously.

\section*{Acknowledgements}

Thank you to Mandy Guo and Gustavo Hernandez Abrego for discussion and initial analyis. We thank Sebastian Ruder and Melvin Johnson for helpful comments on an earlier draft of this paper. We also thank Rattima Nitisaroj for helping us evaluate the quality of our Thai sentence breaking.

\bibliographystyle{acl_natbib}
\bibliography{mainbib}

\begin{thebibliography}{23}
\expandafter\ifx\csname natexlab\endcsname\relax\def\natexlab#1{#1}\fi

\bibitem[{Ahmad et~al.(2019)Ahmad, Constant, Yang, and Cer}]{ReQA}
Amin Ahmad, Noah Constant, Yinfei Yang, and Daniel Cer. 2019.
\newblock \href {https://doi.org/10.18653/v1/D19-5819} {{R}e{QA}: An evaluation
  for end-to-end answer retrieval models}.
\newblock In \emph{Proceedings of the 2nd Workshop on Machine Reading for
  Question Answering}, pages 137--146, Hong Kong, China. Association for
  Computational Linguistics.

\bibitem[{Aroonmanakun(2007)}]{thai-breaking}
Wirote Aroonmanakun. 2007.
\newblock Thoughts on word and sentence segmentation in {Thai}.
\newblock In \emph{Proceedings of the Seventh Symposium on Natural language
  Processing, Pattaya, Thailand, December 13--15}, pages 85--90.

\bibitem[{Artetxe et~al.(2019)Artetxe, Ruder, and Yogatama}]{XQuAD}
Mikel Artetxe, Sebastian Ruder, and Dani Yogatama. 2019.
\newblock On the cross-lingual transferability of monolingual representations.
\newblock \emph{arXiv preprint arXiv:1910.11856}.

\bibitem[{Artetxe and Schwenk(2019)}]{LASER}
Mikel Artetxe and Holger Schwenk. 2019.
\newblock Massively multilingual sentence embeddings for zero-shot
  cross-lingual transfer and beyond.
\newblock \emph{Transactions of the Association for Computational Linguistics},
  7:597--610.

\bibitem[{Clark et~al.(2020)Clark, Choi, Collins, Garrette, Kwiatkowski,
  Nikolaev, and Palomaki}]{TyDiQA}
Jonathan~H. Clark, Eunsol Choi, Michael Collins, Dan Garrette, Tom Kwiatkowski,
  Vitaly Nikolaev, and Jennimaria Palomaki. 2020.
\newblock \href {http://arxiv.org/abs/2003.05002} {{TyDi QA}: A benchmark for
  information-seeking question answering in typologically diverse languages}.

\bibitem[{Conneau et~al.(2019)Conneau, Khandelwal, Goyal, Chaudhary, Wenzek,
  Guzm{\'a}n, Grave, Ott, Zettlemoyer, and Stoyanov}]{XLM-R}
Alexis Conneau, Kartikay Khandelwal, Naman Goyal, Vishrav Chaudhary, Guillaume
  Wenzek, Francisco Guzm{\'a}n, Edouard Grave, Myle Ott, Luke Zettlemoyer, and
  Veselin Stoyanov. 2019.
\newblock Unsupervised cross-lingual representation learning at scale.
\newblock \emph{arXiv preprint arXiv:1911.02116}.

\bibitem[{Conneau and Lample(2019)}]{XLM}
Alexis Conneau and Guillaume Lample. 2019.
\newblock \href
  {http://papers.nips.cc/paper/8928-cross-lingual-language-model-pretraining.pdf}
  {Cross-lingual language model pretraining}.
\newblock In H.~Wallach, H.~Larochelle, A.~Beygelzimer, F.~d\textquotesingle
  Alch\'{e}-Buc, E.~Fox, and R.~Garnett, editors, \emph{Advances in Neural
  Information Processing Systems 32}, pages 7059--7069. Curran Associates, Inc.

\bibitem[{Conneau et~al.(2018)Conneau, Rinott, Lample, Williams, Bowman,
  Schwenk, and Stoyanov}]{XNLI}
Alexis Conneau, Ruty Rinott, Guillaume Lample, Adina Williams, Samuel Bowman,
  Holger Schwenk, and Veselin Stoyanov. 2018.
\newblock {XNLI}: Evaluating cross-lingual sentence representations.
\newblock In \emph{Proceedings of the 2018 Conference on Empirical Methods in
  Natural Language Processing}, pages 2475--2485.

\bibitem[{Devlin et~al.(2019)Devlin, Chang, Lee, and Toutanova}]{BERT}
Jacob Devlin, Ming-Wei Chang, Kenton Lee, and Kristina Toutanova. 2019.
\newblock \href {https://www.aclweb.org/anthology/N19-1423} {{BERT}:
  Pre-training of deep bidirectional transformers for language understanding}.
\newblock In \emph{Proceedings of the 2019 Conference of the North {A}merican
  Chapter of the Association for Computational Linguistics: Human Language
  Technologies, Volume 1 (Long and Short Papers)}, pages 4171--4186,
  Minneapolis, Minnesota. Association for Computational Linguistics.

\bibitem[{Gillick et~al.(2018)Gillick, Presta, and Tomar}]{gillick2018end}
Daniel Gillick, Alessandro Presta, and Gaurav~Singh Tomar. 2018.
\newblock End-to-end retrieval in continuous space.
\newblock \emph{arXiv preprint arXiv:1811.08008}.

\bibitem[{Glava{\v{s}} et~al.(2019)Glava{\v{s}}, Litschko, Ruder, and
  Vuli{\'c}}]{BLI}
Goran Glava{\v{s}}, Robert Litschko, Sebastian Ruder, and Ivan Vuli{\'c}. 2019.
\newblock \href {https://doi.org/10.18653/v1/P19-1070} {How to (properly)
  evaluate cross-lingual word embeddings: On strong baselines, comparative
  analyses, and some misconceptions}.
\newblock In \emph{Proceedings of the 57th Annual Meeting of the Association
  for Computational Linguistics}, pages 710--721, Florence, Italy. Association
  for Computational Linguistics.

\bibitem[{Henderson et~al.(2017)Henderson, Al{-}Rfou, Strope, Sung,
  Luk{\'{a}}cs, Guo, Kumar, Miklos, and Kurzweil}]{hendersonASSLGK17}
Matthew Henderson, Rami Al{-}Rfou, Brian Strope, Yun{-}Hsuan Sung,
  L{\'{a}}szl{\'{o}} Luk{\'{a}}cs, Ruiqi Guo, Sanjiv Kumar, Balint Miklos, and
  Ray Kurzweil. 2017.
\newblock \href {http://arxiv.org/abs/1705.00652} {Efficient natural language
  response suggestion for smart reply}.
\newblock \emph{CoRR}, abs/1705.00652.

\bibitem[{Hu et~al.(2020)Hu, Ruder, Siddhant, Neubig, Firat, and
  Johnson}]{xtreme}
Junjie Hu, Sebastian Ruder, Aditya Siddhant, Graham Neubig, Orhan Firat, and
  Melvin Johnson. 2020.
\newblock {XTREME}: A massively multilingual multi-task benchmark for
  evaluating cross-lingual generalization.
\newblock \emph{arXiv preprint arXiv:2003.11080}.

\bibitem[{Lewis et~al.(2019)Lewis, O{\u{g}}uz, Rinott, Riedel, and
  Schwenk}]{MLQA}
Patrick Lewis, Barlas O{\u{g}}uz, Ruty Rinott, Sebastian Riedel, and Holger
  Schwenk. 2019.
\newblock {MLQA}: Evaluating cross-lingual extractive question answering.
\newblock \emph{arXiv preprint arXiv:1910.07475}.

\bibitem[{Pires et~al.(2019)Pires, Schlinger, and
  Garrette}]{how-multi-is-mbert}
Telmo Pires, Eva Schlinger, and Dan Garrette. 2019.
\newblock How multilingual is multilingual {BERT}?
\newblock In \emph{Proceedings of the 57th Annual Meeting of the Association
  for Computational Linguistics}, pages 4996--5001.

\bibitem[{Rajpurkar et~al.(2016)Rajpurkar, Zhang, Lopyrev, and Liang}]{SQuAD}
Pranav Rajpurkar, Jian Zhang, Konstantin Lopyrev, and Percy Liang. 2016.
\newblock \href {https://doi.org/10.18653/v1/D16-1264} {{SQ}u{AD}: 100,000+
  questions for machine comprehension of text}.
\newblock In \emph{Proceedings of the 2016 Conference on Empirical Methods in
  Natural Language Processing}, pages 2383--2392, Austin, Texas. Association
  for Computational Linguistics.

\bibitem[{Schwenk and Li(2018)}]{MLDoc}
Holger Schwenk and Xian Li. 2018.
\newblock A corpus for multilingual document classification in eight languages.
\newblock In \emph{Proceedings of the Eleventh International Conference on
  Language Resources and Evaluation (LREC 2018)}.

\bibitem[{Siddhant et~al.(2020)Siddhant, Bapna, Tsai, Riesa, Raman, Johnson,
  Ari, and Firat}]{nmt_xlingual}
Aditya Siddhant, Ankur Bapna, Henry Tsai, Jason Riesa, Karthik Raman, Melvin
  Johnson, Naveen Ari, and Orhan Firat. 2020.
\newblock Evaluating the cross-lingual effectiveness of massively multilingual
  neural machine translation.
\newblock \emph{arXiv preprint arXiv:1909.00437}.

\bibitem[{Singh et~al.(2019)Singh, McCann, Keskar, Xiong, and Socher}]{XLDA}
Jasdeep Singh, Bryan McCann, Nitish~Shirish Keskar, Caiming Xiong, and Richard
  Socher. 2019.
\newblock \href {http://arxiv.org/abs/1905.11471} {{XLDA}: Cross-lingual data
  augmentation for natural language inference and question answering}.
\newblock \emph{CoRR}, abs/1905.11471.

\bibitem[{Wu and Dredze(2019)}]{surprising-xlingual-effectiveness}
Shijie Wu and Mark Dredze. 2019.
\newblock {B}eto, {B}entz, {B}ecas: The surprising cross-lingual effectiveness
  of {BERT}.
\newblock In \emph{Proceedings of the 2019 Conference on Empirical Methods in
  Natural Language Processing and the 9th International Joint Conference on
  Natural Language Processing (EMNLP-IJCNLP)}, pages 833--844.

\bibitem[{Yang et~al.(2019{\natexlab{a}})Yang, Cer, Ahmad, Guo, Law, Constant,
  Abrego, Yuan, Tar, Sung, Strope, and Kurzweil}]{USE-QA}
Yinfei Yang, Daniel Cer, Amin Ahmad, Mandy Guo, Jax Law, Noah Constant,
  Gustavo~Hernandez Abrego, Steve Yuan, Chris Tar, Yun-Hsuan Sung, Brian
  Strope, and Ray Kurzweil. 2019{\natexlab{a}}.
\newblock Multilingual universal sentence encoder for semantic retrieval.
\newblock \emph{arXiv preprint arXiv:1907.04307}.

\bibitem[{Yang et~al.(2019{\natexlab{b}})Yang, Zhang, Tar, and
  Baldridge}]{PAWSX}
Yinfei Yang, Yuan Zhang, Chris Tar, and Jason Baldridge. 2019{\natexlab{b}}.
\newblock {PAWS-X}: A cross-lingual adversarial dataset for paraphrase
  identification.
\newblock In \emph{Proceedings of the 2019 Conference on Empirical Methods in
  Natural Language Processing and the 9th International Joint Conference on
  Natural Language Processing (EMNLP-IJCNLP)}, pages 3678--3683.

\bibitem[{Zweigenbaum et~al.(2017)Zweigenbaum, Sharoff, and Rapp}]{BUCC}
Pierre Zweigenbaum, Serge Sharoff, and Reinhard Rapp. 2017.
\newblock Overview of the second {BUCC} shared task: Spotting parallel
  sentences in comparable corpora.
\newblock In \emph{Proceedings of the 10th Workshop on Building and Using
  Comparable Corpora}, pages 60--67.

\end{thebibliography}
\clearpage
\appendix

\section{USE-QA} \label{sec:USE_QA}

We test the Universal Sentence Encoder Multilingual QA \cite{USE-QA}, which specifically targets cross-lingual QA retrieval on our LAReQA benchmark. The architecture and training details of USE-QA are provided in \citet{USE-QA}. We use the USE-QA model out of the box\footnote{\url{https://tfhub.dev/google/universal-sentence-encoder-multilingual-qa}} without fine-tuning on any SQuAD data, as it was already trained specifically for retrieval QA\@. As USE-QA only supports 8 of the 11 XQuAD languages (ar, de, en, es, ru, th, tr, zh), we restrict our evaluation to these languages when comparing USE-QA to other models.
\begin{table}[!htb]
\centering
\setlength\tabcolsep{5pt} 
\begin{tabular}{l|c}
                 & $\text{\xquadr{}}_{USE}$\\
\hline
En-En       &      0.33      \\
X-X         &      0.25   \\
X-X-mono    &      0.55   \\
X-Y          &     0.67   \\
\TranslateEn  &   0.73    \\
USE-QA         &   0.51  \\ 
\end{tabular}
\caption{Mean average precision (mAP) of baseline models on $\text{\xquadr{}}_{USE}$, a version of \xquadr{} restricted to the 8 languages supported by USE-QA.}
\label{table:results_USE_QA}
\end{table}%

From Table~\ref{table:results_USE_QA}, we can see USE-QA is not competitive with the mBERT baselines, despite being trained specifically for QA retrieval over a large in-house QA dataset. However, it may be possible to improve this performance by fine-tuning for SQuAD retrieval.

\section{Language Distributions of Top Results}
\label{sec:top100_bias}

\begin{figure*}[htb!]
\centering
\begin{subfigure}{.35\textwidth}
  \centering
  \includegraphics[width=\linewidth]{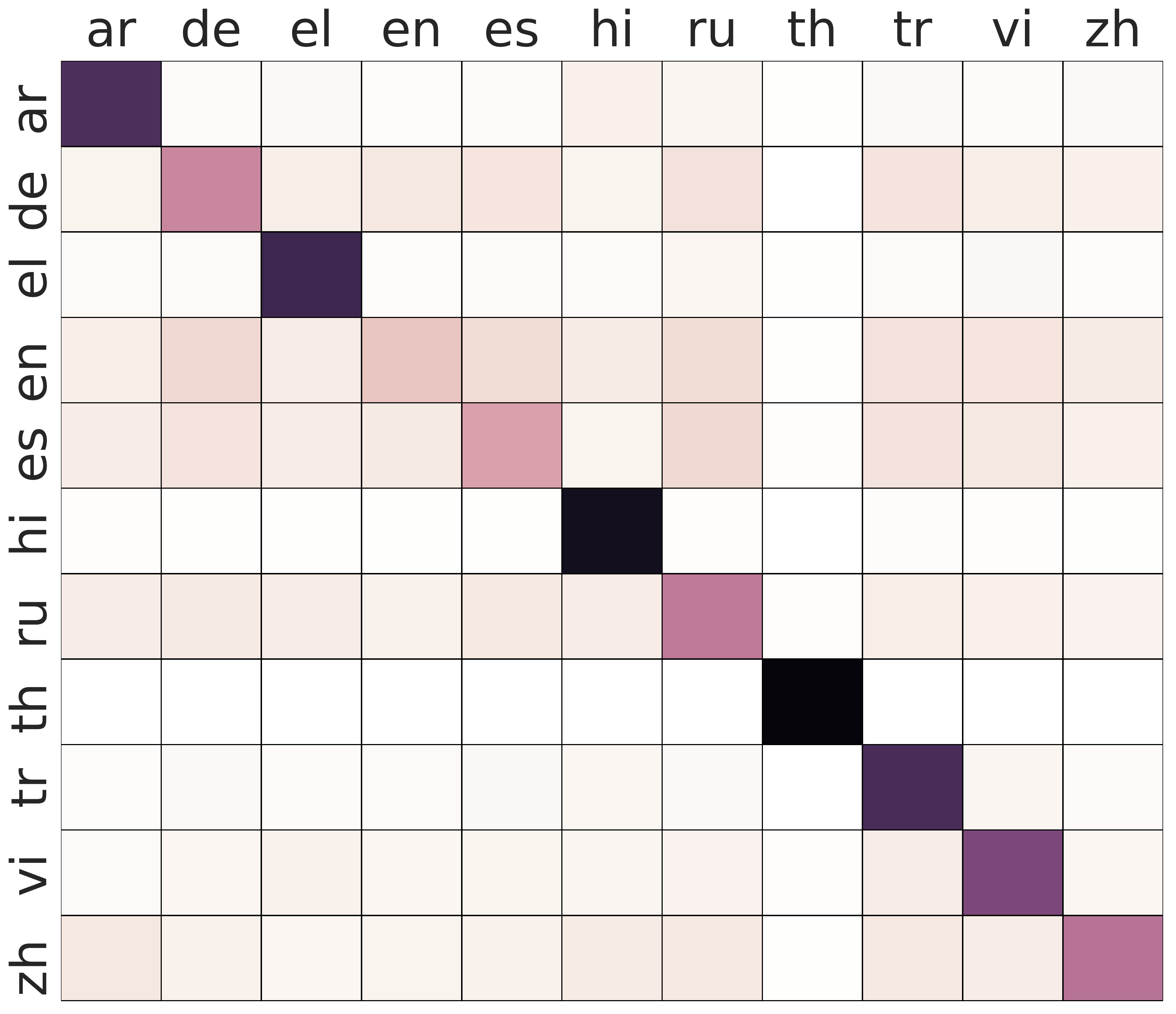}
  \caption{\EnEn{}}
\end{subfigure}%
\begin{subfigure}{.35\textwidth}
  \centering
  \includegraphics[width=\linewidth]{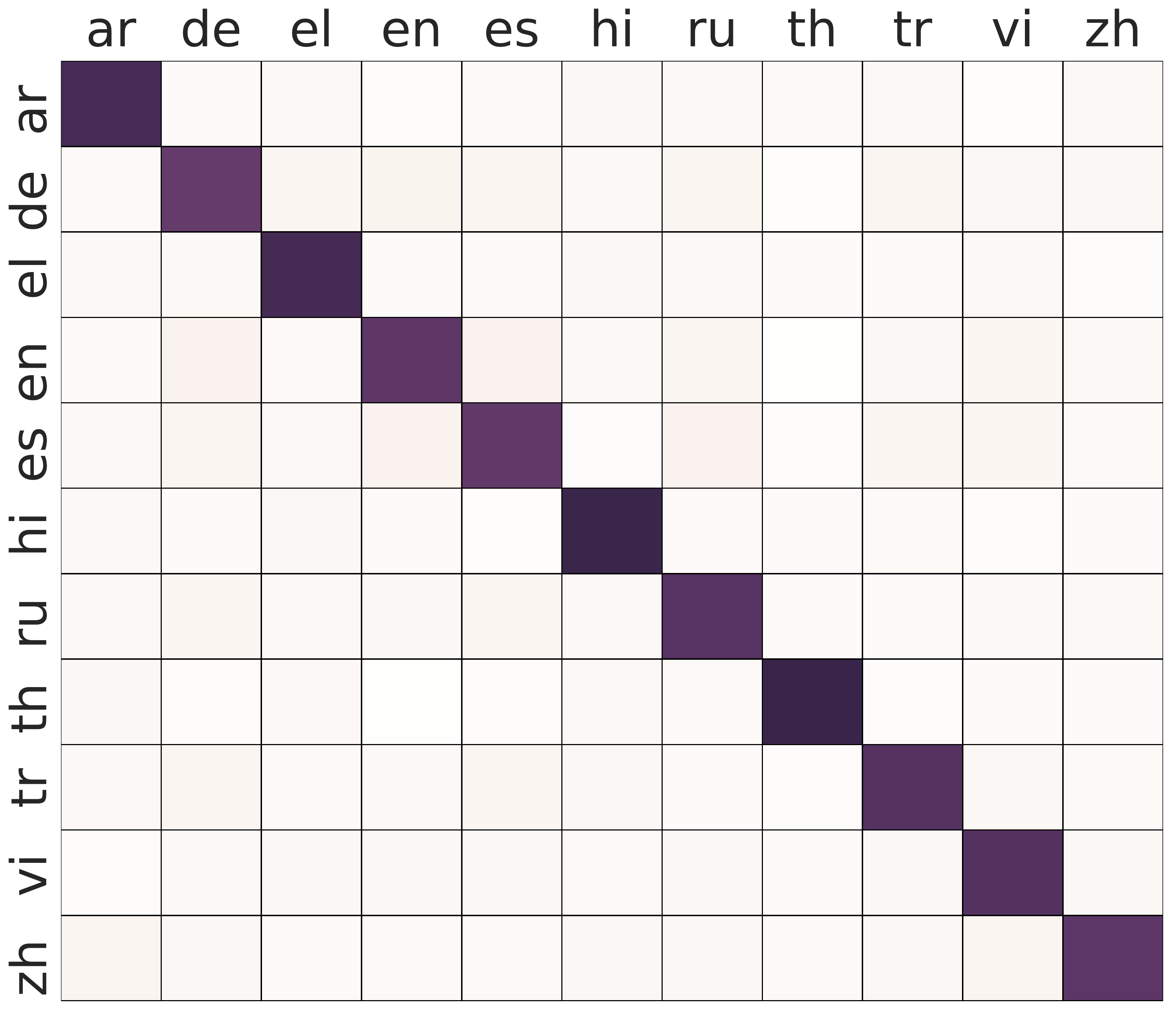}
  \caption{\XX{}}
\end{subfigure}
\begin{subfigure}{.048\textwidth}
  \centering
  \raisebox{12pt}{\includegraphics[width=\linewidth]{final_single_correct_ans/colorbar.pdf}}
\end{subfigure}
\begin{subfigure}{.35\textwidth}
  \centering
  \includegraphics[width=\linewidth]{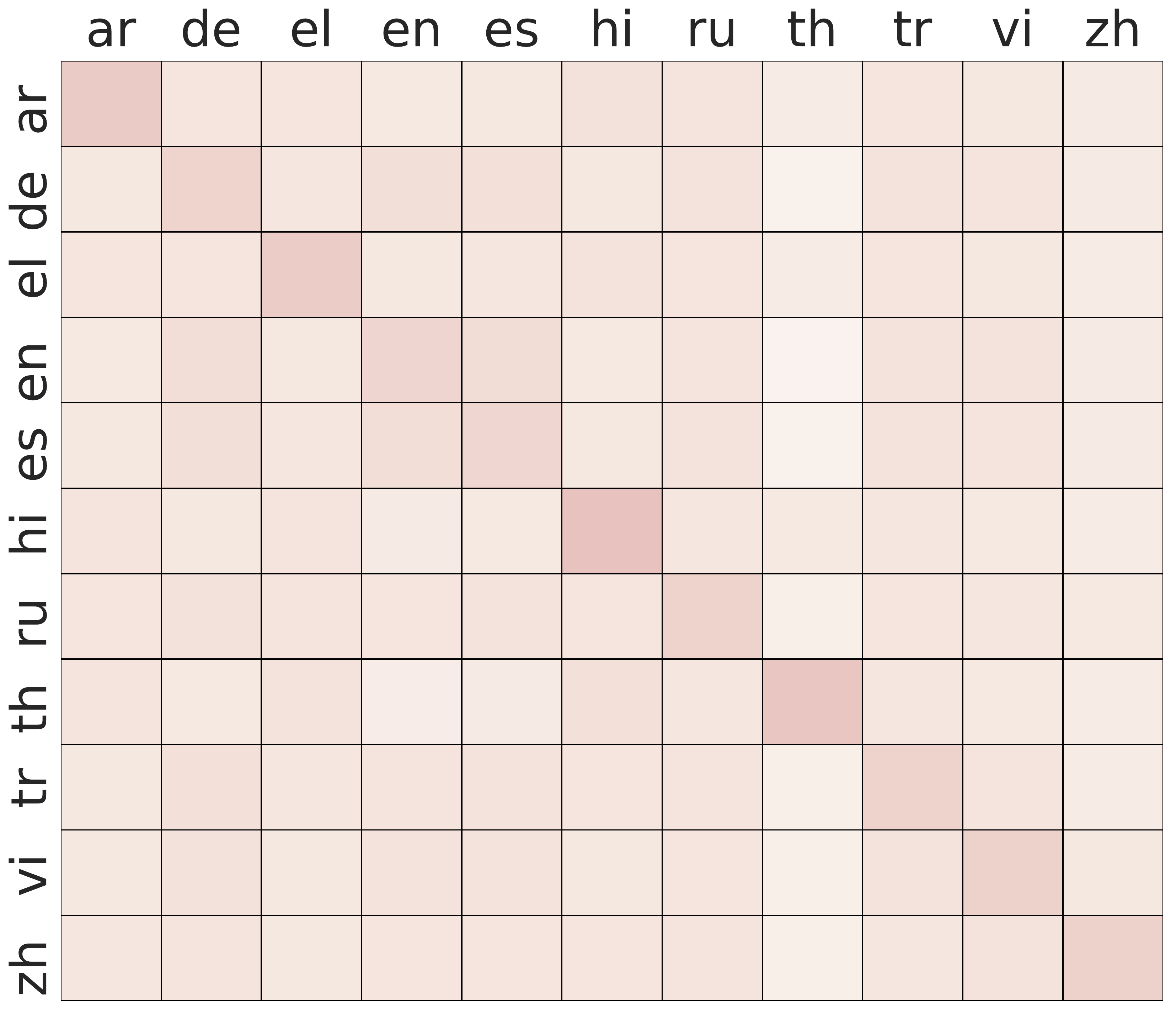}
  \caption{\XXmono{}}
\end{subfigure}%
\begin{subfigure}{.35\textwidth}
  \centering
  \includegraphics[width=\linewidth]{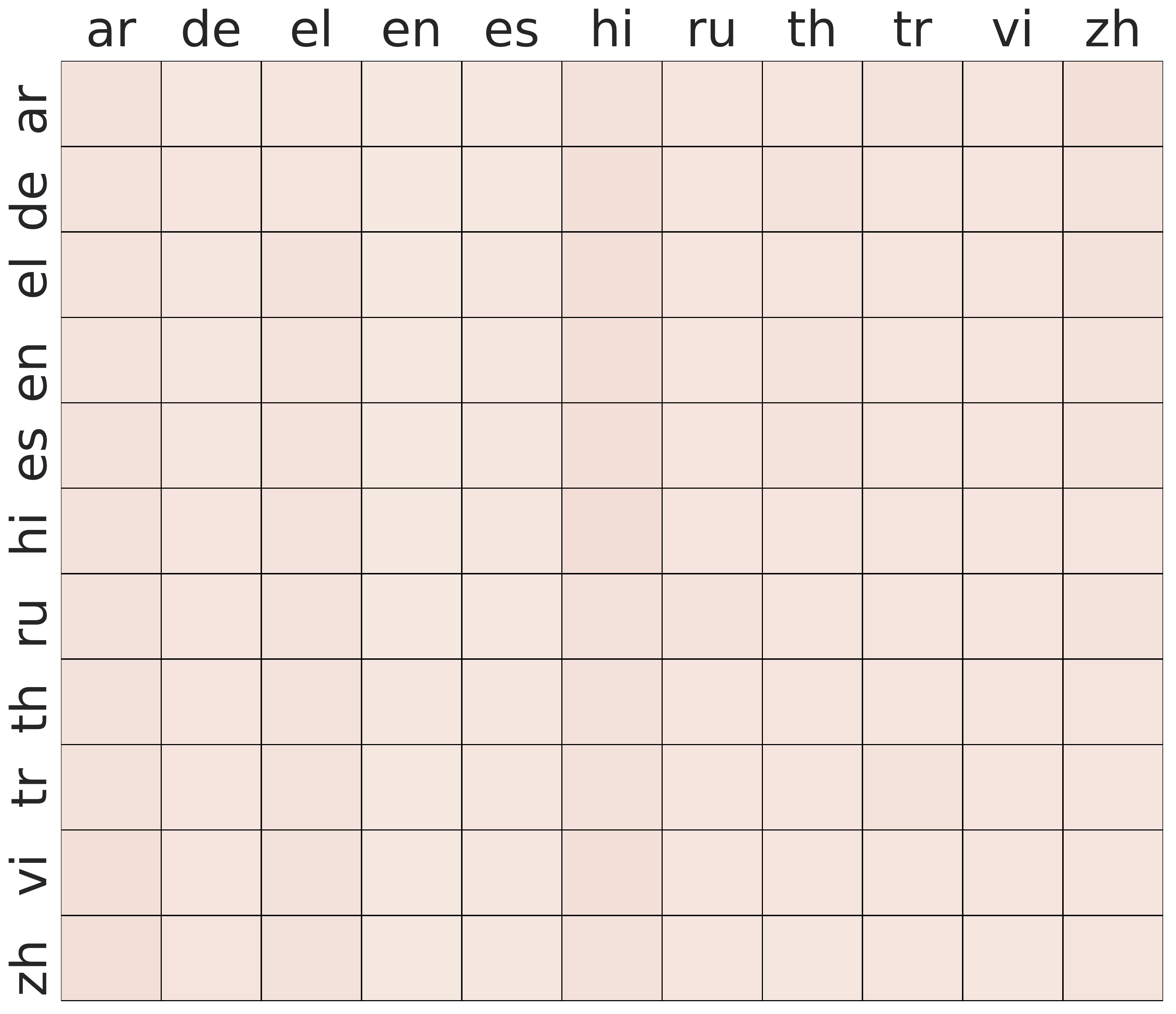}
  \caption{\XY{}}
\end{subfigure}
\begin{subfigure}{.048\textwidth}
  \centering
  \raisebox{12pt}{\includegraphics[width=\linewidth]{final_single_correct_ans/colorbar.pdf}}
\end{subfigure}
\caption{Proportion of top-100 retrieved answers that are in a given language (column), broken down by question language (row). Each row sums to 1.0.}
\label{fig:top100}
\end{figure*}

Our core LAReQA mAP metric tests for both question answering (QA) matching ability, as well as absence of language bias. We can factor out QA performance and focus more directly on language bias by simply ignoring which answers are correct, and observing the distribution of languages that a model retrieves among its top-ranked candidates.

The heatmaps in Figure~\ref{fig:top100} show for each question language (row), the frequency of different answer languages (column) among the top 100 retrieved candidates, for each of our baseline models on the \xquadr{} dataset. The strong diagonal in \XX{} indicates that when the question is in a given language, nearly all of the top 100 retrieved results are in the same language. Overall, this measure of language bias is consistent with those discussed in Section~\ref{sec:lang-bias}, with the models ranking \XY{}\,$>$\,\XXmono{}\,$>$\,\EnEn{}\,$>$\,\XX{}.

Interestingly, \XY{} performs almost perfectly on this ``semantics-free'' measurement of language bias. This is in contrast to the mAP performance of the same model in Figure~\ref{fig:XY_single_ans}, where the retrieval of \emph{correct} answers is somewhat improved when the Q and A languages match. Taken together, we can say that \XY{} is nearly perfectly unbiased in which languages it retrieves \emph{on the whole}, but is slightly biased as to which language pairs exhibit the strongest QA matching.

\end{document}